\documentclass{article}
\usepackage[preprint]{neurips_2026}
\makeatletter
\renewcommand{\@noticestring}{Preprint. Accepted at the 40th Conference on Neural Information Processing Systems (NeurIPS 2026).}
\makeatother

\usepackage[utf8]{inputenc} 
\usepackage[T1]{fontenc}    
\usepackage{url}            
\usepackage{booktabs}       
\usepackage{amsfonts}       
\usepackage{nicefrac}       
\usepackage{microtype}      
\usepackage{xcolor}         
\usepackage{multirow}       

\usepackage{graphicx}
\usepackage{amsmath}
\usepackage{algorithm}
\usepackage{algorithmic}
\usepackage{float}

\title{Aligning LLMs with Biomedical Knowledge using Balanced Fine-Tuning}

\author{%
\textbf{Zhenchao Tang$^{*1}$, Fang Wang$^{*\dagger1}$, Haohuai He$^{1}$, Jiale Zhou$^{1}$} \\
\textbf{Tianxu Lv$^{1}$, Jun Zhu$^{1}$, Shouzhi Chen$^{1}$, Minghao Yang$^{1}$} \\
\textbf{Yu Wang$^{1}$, Jiayang Wu$^{1}$, Yidong Song$^{2}$, Yaokun Li$^{3}$} \\
\textbf{Jiehui Huang$^{4}$, Jun Zhou$^{5}$, Bing He$^{1}$, Jianhua Yao$^{\dagger1}$} \\
$^{1}$Tencent AI for Life Sciences Lab \\
$^{2}$Nankai University; $^{3}$Sun Yat-sen University\\
$^{4}$The Hong Kong University of Science and Technology\\
$^{5}$Tsinghua University\\
$^{*}$Equal contribution\\
$^{\dagger}$Corresponding authors: avonwanghit@gmail.com, jianhua.yao@gmail.com
}

\begin{document}
\raggedbottom 

\maketitle

\begin{abstract}
Engineering LLMs to accelerate life sciences research requires a robust alignment with biomedical knowledge. We observe that biomedical text exhibits a fundamentally different uncertainty structure from general text: dense low-confidence runs encode epistemic knowledge gaps (dense causal chains, rare entities) rather than the sparse aleatoric stylistic variation typical of general text. Based on this discovery, we propose Balanced Fine-Tuning (BFT), a dual-scale post-training method that combines group-normalized token reweighting with sequence-level reallocation toward knowledge-dense samples exhibiting dense epistemic uncertainty. Across medical evaluation, biological reasoning, sparse-reward RL, and biological representation tasks, BFT provides more consistent gains than SFT and DFT under a shared training setup. When replacing the default closed-source backbones in GeneAgent (GPT-4o) and VCWorld (Gemini-2.5-Flash), the BFT-aligned 70B model delivers stronger performance across biological process reasoning and chemical perturbation prediction. Critically, all BFT variants further improve after subsequent GRPO with sparse rewards, while SFT and DFT degrade, suggesting that epistemic-aware post-training provides a more robust policy initialization. Beyond text generation, BFT-aligned LLMs produce more accurate and professional biomedical profile texts; after encoding these profiles with a text embedding model, the resulting representations support gene-level, cell-level, and perturbation-response tasks, suggesting that BFT-enhanced generation can facilitate biological representation and, in turn, broader biomedical downstream tasks.
\end{abstract}

\section{Introduction}

\begin{figure}[t]
\centering
\includegraphics[width=\linewidth]{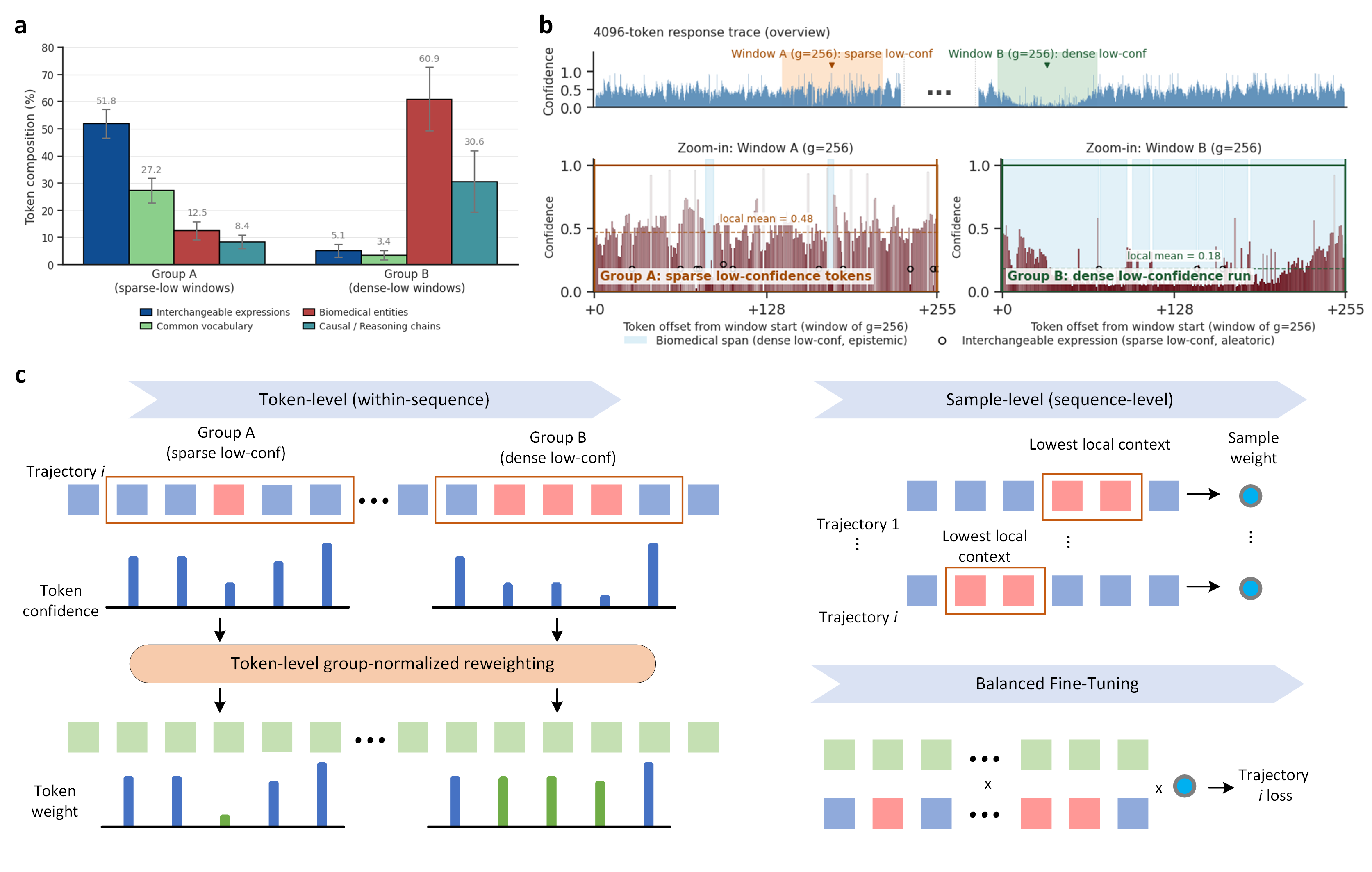}
\caption{Overview of BFT. \textbf{a}: Semantic composition of low-confidence tokens inside Group~A (sparse-low) and Group~B (dense-low). \textbf{b}: 4096-token trace illustrating one sparse-low window (Group~A) and one dense-low window (Group~B). \textbf{c}: BFT applies token- and sample-level adjustments.}
\label{Figure1}
\end{figure}

Post-training is the practical bridge for aligning Large Language Models (LLMs) with specialized biomedical knowledge. Standard strategies such as Supervised Fine-Tuning (SFT) and Reinforcement Learning (RL) have driven rapid progress in general domains \cite{guo2025deepseek}, but confidence-based reweighting creates a domain-specific risk: a token the model finds hard to predict need not carry little semantic weight. Low probability can reflect interchangeable surface forms, but in biomedical text low-confidence regions often contain rare entity names, pathway nodes, and mechanistic steps whose omission changes the passage. We use \emph{aleatoric} and \emph{epistemic} as interpretations of these two regimes, not as directly observed labels; the diagnostic below instead measures their spatial structure and external concept enrichment. Dynamic Fine-Tuning (DFT) \cite{wu2025generalization} stabilizes SFT by scaling each token's loss by its predicted probability, but an absolute-confidence rule cannot distinguish an isolated outlier from a whole region the model finds difficult. RL faces a complementary obstacle: biomedical rewards are typically sparse and available only at the outcome level, such as whether a predicted gene-response category matches the observed perturbation outcome \cite{qi2024large, liu2025application, qu2025crispr}, so the initialization must already place the model in a useful reasoning region.

To ground this empirically, we ran a diagnostic teacher-forcing pass over PubMed Central open-access full text \cite{pmc_oa}. Because a diagnostic should describe the setting a fine-tuning objective actually operates in, each passage is cast as an instruction--response pair, with the article title and section heading as the prompt $x$ and the section body as the reference response $y^*$. Unless otherwise noted, the diagnostic uses the pre-fine-tuning DeepSeek-R1-Distill-Llama-70B model. At each reference token we record $c_t := \pi_\theta(y_t^* \mid y_{<t}^*, x)$, slide a window of size $g=256$ along the confidence sequence, and split the windows by the spatial structure of their low-confidence tokens: \emph{Group~A} holds those whose low-confidence tokens are sparsely scattered, \emph{Group~B} those whose low-confidence tokens form a long contiguous run. Exact cutoffs are given in Appendix~\ref{app:diag}; they describe the two regimes and never enter the BFT objective, which depends only on continuous quantities.

Tagging low-confidence tokens with SciSpacy \cite{neumann-etal-2019-scispacy} and a causal-connective list, Group~A is dominated by interchangeable expressions and Group~B by biomedical entities and causal chains (Figure~\ref{Figure1}a, with a single-response trace in Figure~\ref{Figure1}b). We test this interpretation with three complementary controls. \textbf{First, a within-corpus and cross-domain control:} the entity share among low-confidence tokens rises from 12.5\% in Group~A to 60.9\% in Group~B on PubMed Central and from 11.8\% to 57.4\% on our synthetic supervision, but only from 3.6\% to 4.1\% on UltraChat-200k \cite{ding2023enhancing} and from 3.0\% to 3.2\% on NuminaMath-CoT \cite{numina_math_datasets}, the last despite no shortage of hard windows. Because both terms of each contrast share one corpus and one annotator, this comparison reduces the influence of overall corpus-specific annotation rates, although differential annotation errors may remain; the similar shifts on real and synthetic biomedical text further argue against a supervision-generation artifact. \textbf{Second, a human-annotation control:} on CRAFT \cite{bada2012concept}, an expert-annotated corpus drawn from the same source, 63.8\% of dense-window low-confidence tokens fall inside manually marked concept spans, against 11.2\% in sparse windows. \textbf{Third, a retention counterfactual:} dense-only reaches 0.674 on VCWorld, versus 0.571 for a size-matched random set, 0.508 for sparse-only, and 0.562 for SFT---an ordering inconsistent with a pure ``retain more tokens'' explanation (Appendix~\ref{app:diag}).

Building on this analysis, we propose Balanced Fine-Tuning (BFT), which reconciles optimization stability with adaptive learning from under-confident, knowledge-dense biomedical samples (Figure~\ref{Figure1}c). BFT acts at two scales through a single sliding window of size $g$. At the token level it replaces DFT's absolute-confidence weight with the token's confidence \emph{relative} to its local context, so that a low-confidence token inside an otherwise confident window is treated as an outlier while one inside an equally uncertain window retains its learning signal. At the sample level it reuses the same local confidences, taking their minimum as a bounded hard-sample coefficient $s_b$ that reallocates budget across sequences. The two are complementary: the token weight allocates budget \emph{within} a sequence by relative local difficulty, $s_b$ allocates it \emph{across} sequences by absolute hardness. Both signals are detached by stop-gradient, and $g$ is the only additional tunable train-time hyperparameter.

\begin{figure}[t]
\centering
\includegraphics[width=\linewidth]{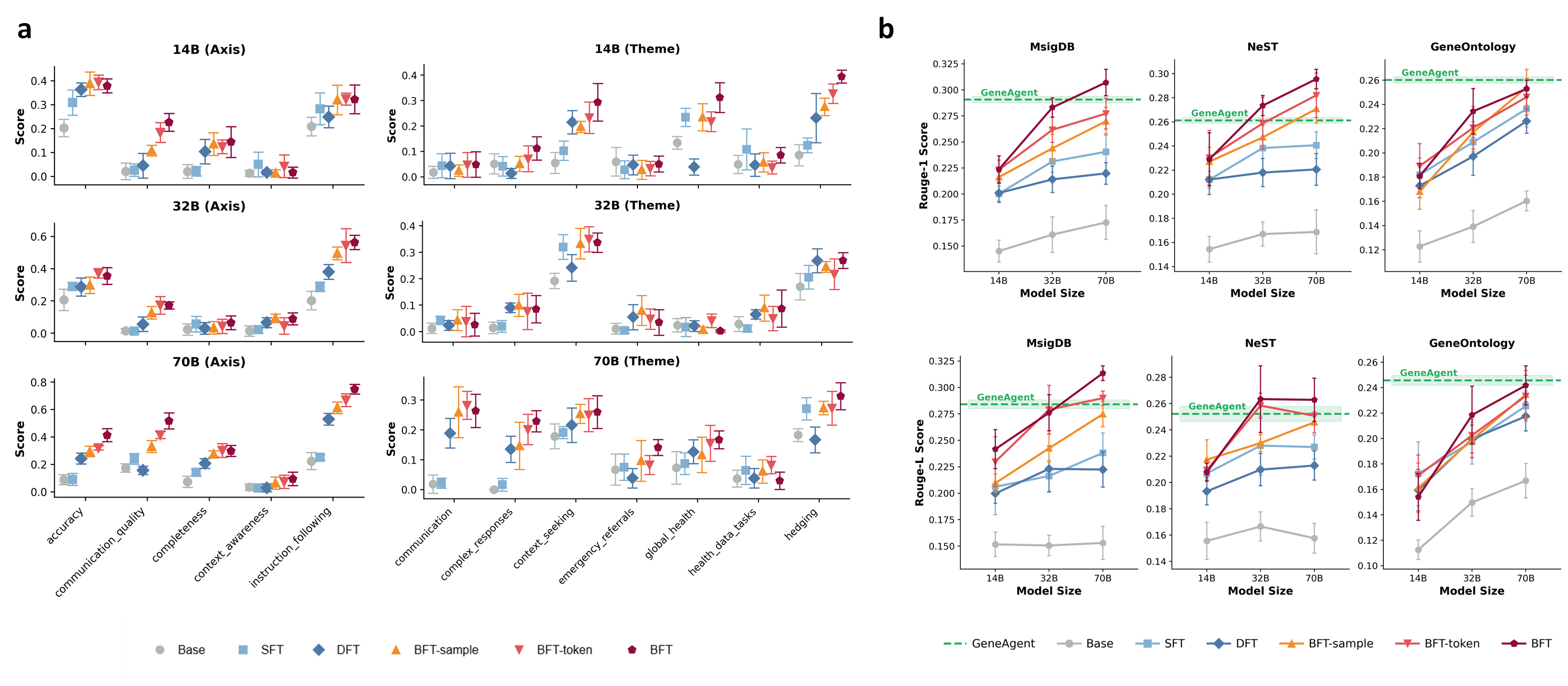}
\caption{
Medical competence and biological process reasoning. \textbf{a}: Medical competence is measured across axis-wise (left) and theme-wise (right). \textbf{b}: Biological reasoning benchmark.
}
\label{Figure2}
\end{figure}

The effect of BFT extends beyond text generation. Computational biology is split between generative tasks that rely on LLMs and discriminative tasks that depend on specialized biological foundation models \cite{tang2025modal}. Since SFT tends to produce generic profiles that blur distinctions between related entities while BFT articulates more precise mechanisms, we ask whether better post-training can narrow this divide through the quality of the generated profile texts themselves.

Our contributions are as follows. \textbf{(1)} We show, under annotator, corpus, length, rarity and counterfactual controls, that dense low-confidence regions in biomedical text are strongly enriched for expert-annotated concepts, and turn this observation into a fine-tuning objective. \textbf{(2)} BFT yields a stronger open-weight 70B biomedical backbone than SFT and DFT across representative agent pipelines (GeneAgent \cite{wang2025geneagent}, VCWorld \cite{yu2025vcworld}); on VCWorld it is not statistically distinguishable from the default closed-source backbone. \textbf{(3)} Under a shared GRPO grid with a sparse task reward and a small format-validity term, all BFT variants improve after subsequent GRPO \cite{shao2024deepseekmath}, whereas SFT and DFT degrade. \textbf{(4)} BFT-generated profile texts, once embedded, yield representations that perform strongly on gene-level prediction, cell-level clustering, multimodal integration and perturbation response prediction.

\section{Related Work}\label{sec:related}

\textbf{Supervised Fine-Tuning} LLM post-training includes supervised fine-tuning (SFT), on-policy distillation \cite{li2026rethinking,zhao2026self,shenfeld2026self}, and reinforcement learning \cite{zhang2025survey,liu2025reinforcement}. Biomedical applications face limited expert supervision and sparse outcome rewards, motivating better use of supervised data. Dynamic Fine-Tuning (DFT) \cite{wu2025generalization} weights token losses by prediction confidence to improve stability and generalization. However, absolute-confidence weighting can suppress rare entities and mechanistic steps within difficult biomedical passages. The same caution applies when uncertain content is not one regime: primitive memory distillation separates recoverable shared semantics from uncertain modality-specific detail \cite{zhang2026modality}, whereas BFT separates sparse and dense low-confidence runs in biomedical text. Other objectives use importance weights, reference-policy anchoring, fitness estimates, or reward-based filtering \cite{qin2025supervised,liu2025uft,zhu2025anchored,liu2026profit,dong2023raft}. BFT instead combines local relative-confidence weighting with sequence-level emphasis on hard regions; Appendix~\ref{app:baselines} compares supervised reweighting baselines under the shared recipe. External memory compilers first resolve the current relations and only then choose which records the model reads \cite{song2026contract}; BFT does not select an external store, and its local confidence instead decides which tokens inside the supervised response keep their learning signal. A later GRPO stage need not recover a signal left implicit by supervised training \cite{zhu2026comet}. Local relationships have also been exploited in weakly supervised visual posture mining \cite{tang2023weakly}; that work models relations among image regions, whereas BFT normalizes token confidence within local text contexts. Fixed-size cross-shot memory is likewise gated at visual boundaries so that identity persists without a growing bank \cite{huang2026unityshots}; the BFT window does not store cross-example memory.

\textbf{Biomedical Representations} Specialized representations support tasks from drug--drug interaction prediction \cite{tang2023dsil} to cell annotation and perturbation modeling \cite{hao2024large}. At the molecular level, DMFF-DTA integrates sequence and binding-site information for drug--target affinity prediction \cite{he2025dual}, while DiConSite addresses protein binding-site prediction across ligand modalities \cite{chen2026diconsite}. At the cellular level, scGPT \cite{cui2024scgpt} and Geneformer \cite{theodoris2023transfer} learn from large-scale transcriptomic measurements; scProtoTransformer targets reference mapping across molecules, cells, and donors \cite{tang2026scprototransformer}, scTransMIL links cell profiles to patient-level disease states \cite{tang2026sctransmil}, and scKAN emphasizes interpretable cell-type-specific gene discovery \cite{he2025sckan}. Multimodal approaches such as comprehensive view embedding \cite{tang2024comprehensive} and the modal-nexus auto-encoder \cite{tang2024modal} explicitly integrate heterogeneous cellular measurements. Transformer binding learning likewise separates modality-specific from modality-invariant features before fusion, but for sentiment analysis rather than cellular measurements \cite{huang2024tmbl}. These methods address distinct biological prediction or representation problems rather than biomedical LLM post-training. GenePT \cite{chen2025simple} offers a complementary text-based approach by embedding NCBI gene summaries with \textsc{text-embedding-ada-002} \cite{neelakantan2022text}. Generated content can also look right while dropping concepts required for later retrieval \cite{dong2026images}. We ask whether BFT-generated profiles, encoded with the same text encoder, provide useful gene- and cell-level representations.

\section{Methods}

We briefly recall two prerequisites that BFT builds upon and that are reviewed formally in Appendix~\ref{t-1}: (i) Supervised Fine-Tuning (SFT) is the standard maximum-likelihood objective $-\sum_t \log \pi_\theta(y_t^*\mid y_{<t}^*, x)$, which gives every token unit weight; and (ii) Dynamic Fine-Tuning (DFT) \cite{wu2025generalization} stabilizes SFT by multiplying each token loss by its detached prediction probability $\mathrm{sg}(\pi_\theta(y_t^*\mid y_{<t}^*, x))$, so very low-confidence tokens contribute proportionally less to the gradient. DFT therefore suppresses low-confidence tokens regardless of whether they are isolated outliers or part of a difficult, knowledge-bearing context.

BFT replaces DFT's absolute-confidence rule with a group-normalized one. Dense low-confidence runs in biomedical text often correspond to rare entities, pathway nodes, mutation names, and mechanistic reasoning steps; DFT would suppress these in the same way as stylistic noise. BFT instead asks whether a token is unusually low-confidence \emph{relative to its local biomedical context}: a low-confidence token whose surrounding $g$-token window has high mean confidence (a sparse-low context) is treated as a local outlier, whereas tokens whose surrounding $g$-token window itself has low mean confidence (a dense-low context) retain learning signal because the whole local context is difficult. At the sample level, BFT further reuses the same local context confidences to identify the weakest region of each sequence and turns its minimum value into a bounded hard-sample signal $s_b$.

\paragraph{Per-token confidence}
For each token $t$ of sample $b$, given logits $\mathbf{z}_{b,t} \in \mathbb{R}^V$ ($V$ is the vocabulary size) and target token $y_{b,t}^*$, we define token confidence and token cross-entropy as:
\begin{equation}
c_{b,t}
:= \pi_\theta(y_{b,t}^* \mid y_{b,<t}^*, x_b)
:= \mathrm{softmax}(\mathbf{z}_{b,t})[y_{b,t}^*],
\qquad
\ell_{b,t}
:= -\log c_{b,t}.
\end{equation}

\paragraph{Local context confidence}
BFT uses a single group size $g$, its only additional tunable train-time hyperparameter. Group-level confidence has also been used by DeepConf \cite{fu2025deep} to measure localized reasoning difficulty; here, the local context is used only to quantify biomedical difficulty during training. We define $W_{b,t}$ as a centered window containing at most $g$ valid target positions. Responses shorter than $g$ use the full response; otherwise the window contains token $t$, up to $\lfloor g/2 \rfloor$ left neighbors and up to $\lceil g/2 \rceil-1$ right neighbors, clipped at sequence boundaries:
\begin{equation}
\begin{aligned}
W_{b,t}
&:=
\begin{cases}
\{1,\ldots,|y_b^*|\}, & |y_b^*|<g,\\
\{j:\max(1,t-\lfloor g/2\rfloor)\le j\le
\min(|y_b^*|,t+\lceil g/2\rceil-1)\}, & |y_b^*|\ge g,
\end{cases}\\
C_{b,t}^{\mathrm{loc}}
&:=
\frac{1}{|W_{b,t}|}
\sum_{j \in W_{b,t}} c_{b,j}.
\end{aligned}
\end{equation}
For the default $g=256$, an interior token uses itself, $128$ left neighbors and $127$ right neighbors. Boundary windows contain fewer positions, and responses shorter than $g$ use all their valid response tokens. The local means can be computed efficiently with cumulative sums and normalization by $|W_{b,t}|$.

\paragraph{Token-level group-normalized reweighting}
DFT uses the absolute confidence $c_{b,t}$ as the token weight. BFT instead compares each token with its local context, using the simplest possible group-normalized rule:
\begin{equation}
\rho_{b,t}
:=
\mathrm{clip}\!\left(
\frac{c_{b,t}}{C_{b,t}^{\mathrm{loc}}+\varepsilon},
0,1
\right),
\qquad
w_{b,t}^{\mathrm{BFT}}
:=
\mathrm{sg}\!\left(\rho_{b,t}\right).
\end{equation}
The ratio $\rho_{b,t}\in[0,1]$ measures whether token $t$ is unusually uncertain relative to nearby tokens, and is used directly as the token weight without any further nonlinearity, so BFT introduces no additional hyperparameter beyond the window size $g$. Since $C_{b,t}^{\mathrm{loc}}\le 1$, we have $\rho_{b,t}\ge c_{b,t}/(1+\varepsilon)$; with $\varepsilon=10^{-8}$, the BFT weight is no smaller than the DFT weight up to this numerical tolerance. The difference is governed by local-context uncertainty. In Group~A, $c_{b,t}$ is low while $C_{b,t}^{\mathrm{loc}}$ remains high, so $\rho_{b,t}$ stays close to $c_{b,t}$ and the token is suppressed much as it would be under DFT --- and far below its unit weight under SFT. In Group~B, $C_{b,t}^{\mathrm{loc}}$ is itself low, so $\rho_{b,t}$ moves toward one and the token retains substantially more signal than DFT would give it at the same absolute confidence. For a token with $c_{b,t}=0.10$, a sparse-low context ($C^{\mathrm{loc}}_{b,t}=0.87$) gives $\rho_{b,t}=0.115$ while a dense-low context ($C^{\mathrm{loc}}_{b,t}=0.20$) gives $\rho_{b,t}=0.50$. BFT is thus best read not as suppressing Group~A more aggressively, but as selectively restoring the signal DFT would discard inside knowledge-dense runs.

\paragraph{Why sample-level reweighting is still needed}
Cross-entropy assigns larger losses to low-confidence targets, while $\rho_{b,t}$ redistributes token contributions according to local context. This does not explicitly prioritize entire sequences containing difficult regions. We therefore introduce a sample-level coefficient based on the lowest local confidence. The token-weight comparison with DFT above does not imply that the full product $s_b\rho_{b,t}$ is always larger than the DFT weight.

\paragraph{Sample-level adaptive hard-example weighting}
Concretely, we reuse the same local context confidences $\{C_{b,t}^{\mathrm{loc}}\}_t$ and define the lowest local confidence of each sequence as
\begin{equation}
p_b^{\mathrm{conf}}
:=
\min_{1 \le t \le |y_b^*|}
C_{b,t}^{\mathrm{loc}},
\end{equation}
which identifies the weakest boundary-aware local region in the sequence. The sample-level balance coefficient is
\begin{equation}
s_b
:= \mathrm{sg}\!\big(1 - p_b^{\mathrm{conf}}\big),
\end{equation}
so $s_b \approx 0$ for confident samples and $s_b \approx 1$ for difficult ones, while remaining strictly bounded in $[0,1]$. Token-level and sample-level signals are thus complementary: $w_{b,t}^{\mathrm{BFT}}$ allocates budget \emph{within} a sequence based on relative local difficulty, while $s_b$ allocates budget \emph{across} sequences based on absolute biomedical hardness.

\paragraph{BFT objective}
BFT combines group-normalized token reweighting with sample-level adaptive hard-example learning:
\begin{equation}\label{eq:bft_objective}
\mathcal{L}_{\mathrm{BFT}}(\theta)
=
\frac{1}{B}
\sum_{b=1}^{B}
s_b \,
\frac{
\sum_{t} m_{b,t} \, w_{b,t}^{\mathrm{BFT}} \, \ell_{b,t}
}{
\sum_{t} m_{b,t} + \varepsilon
},
\end{equation}
where $B$ is the mini-batch size, $m_{b,t}=1$ on response target tokens and $0$ on prompt or padding positions, and $\varepsilon$ is a small constant for numerical stability. All confidence-derived weights ($w_{b,t}^{\mathrm{BFT}}$ and $s_b$) are detached from the computation graph. BFT thereby reduces the contribution of sparse low-confidence outliers while preserving stronger optimization signals for difficult, knowledge-dense biomedical runs.

\begin{table}[t]
\centering
\caption{Chemical perturbation reasoning accuracy for joint Up/Down/No classification. Upper block: original VCWorld; middle: aligned 70B replacements; lower: after GRPO. Cells are mean $\pm$ SD over eight stochastic inference passes of one checkpoint (decoding, not training, variability). Displayed averages are rounded to two decimals; tests use unrounded cell-line means. Best in each column is shown in \textbf{bold}.}
\label{tab:vcworld}
\resizebox{\linewidth}{!}{%
\begin{tabular}{l ccccc c}
\toprule
\textbf{Method} & \textbf{C32} & \textbf{HepG2C3A} & \textbf{HOP62} & \textbf{Hs 766T} & \textbf{PANC-1} & \textbf{Average} \\
\midrule
VCWorld (Gemini-2.5-Flash) & 0.69 $\pm$ .01 & 0.67 $\pm$ .02 & 0.72 $\pm$ .01 & 0.69 $\pm$ .02 & 0.62 $\pm$ .02 & 0.68 \\
VCWorld (Base 70B) & 0.42 $\pm$ .02 & 0.37 $\pm$ .02 & 0.41 $\pm$ .02 & 0.39 $\pm$ .02 & 0.40 $\pm$ .02 & 0.40 \\
\midrule
DFT 70B & 0.54 $\pm$ .03 & 0.46 $\pm$ .02 & 0.49 $\pm$ .02 & 0.45 $\pm$ .03 & 0.51 $\pm$ .02 & 0.49 \\
SFT 70B & 0.58 $\pm$ .02 & 0.53 $\pm$ .02 & 0.58 $\pm$ .02 & 0.55 $\pm$ .01 & 0.57 $\pm$ .02 & 0.56 \\
BFT-sample 70B & 0.61 $\pm$ .02 & 0.58 $\pm$ .02 & 0.62 $\pm$ .02 & 0.58 $\pm$ .02 & 0.59 $\pm$ .02 & 0.60 \\
BFT-token 70B & 0.68 $\pm$ .02 & 0.65 $\pm$ .02 & 0.67 $\pm$ .01 & 0.66 $\pm$ .02 & 0.64 $\pm$ .02 & 0.66 \\
BFT 70B & 0.72 $\pm$ .01 & 0.70 $\pm$ .01 & 0.71 $\pm$ .02 & 0.70 $\pm$ .01 & 0.65 $\pm$ .01 & 0.70 \\
\midrule
DFT 70B + GRPO & 0.46 $\pm$ .02 & 0.39 $\pm$ .02 & 0.43 $\pm$ .03 & 0.38 $\pm$ .02 & 0.44 $\pm$ .02 & 0.42 ($\downarrow$ 0.07) \\
SFT 70B + GRPO & 0.51 $\pm$ .02 & 0.45 $\pm$ .02 & 0.51 $\pm$ .02 & 0.46 $\pm$ .02 & 0.50 $\pm$ .02 & 0.49 ($\downarrow$ 0.08) \\
BFT-sample 70B + GRPO & 0.65 $\pm$ .02 & 0.59 $\pm$ .02 & 0.65 $\pm$ .02 & 0.62 $\pm$ .02 & 0.61 $\pm$ .02 & 0.62 ($\uparrow$ 0.03) \\
BFT-token 70B + GRPO & 0.72 $\pm$ .02 & 0.70 $\pm$ .02 & 0.71 $\pm$ .02 & 0.69 $\pm$ .02 & 0.65 $\pm$ .02 & 0.69 ($\uparrow$ 0.03) \\
BFT 70B + GRPO & \textbf{0.76} $\pm$ .01 & \textbf{0.74} $\pm$ .01 & \textbf{0.75} $\pm$ .01 & \textbf{0.72} $\pm$ .01 & \textbf{0.71} $\pm$ .01 & \textbf{0.74} ($\uparrow$ 0.04) \\
\bottomrule
\end{tabular}%
}
\end{table}

\section{Experimental Setup}

\subsection{Datasets and Tasks}

We use two complementary sources for joint post-training. \textbf{Medical trajectory synthesis.} Each sample of the OpenAI Health Bench~\cite{arora2025healthbench} \textit{Consensus} subset pairs a clinical instruction with a set of rubrics covering factual accuracy, completeness, reasoning depth and safety. We synthesize response trajectories by giving GPT-5.1 the instruction together with its full rubric set and prompting it to satisfy the rubric dimensions jointly. \textbf{Biological knowledge synthesis.} Following GenePT \cite{chen2025simple}, the NCBI-derived biomedical corpus is used as the knowledge source in EasyDataset \cite{miao2025easy}, and GPT-OSS-120B \cite{agarwal2025gpt} synthesizes instruction--response pairs covering broad biological knowledge. Because \textit{Consensus} and the \textit{Hard} subset used for evaluation overlap, we remove the intersection by prompt id before generating any supervision; Appendix~\ref{app:data} details this operation, the subsequent $n$-gram decontamination, filtering rules and mixture construction.

We evaluate \textbf{medical competence} on HealthBench \textit{Hard}, reporting theme- and axis-wise scores \cite{arora2025healthbench}. For \textbf{biological process reasoning}, we replace GeneAgent's GPT-4o backbone while retaining its pipeline \cite{wang2025geneagent}. For \textbf{chemical perturbation reasoning}, we replace VCWorld's Gemini-2.5-Flash backbone and predict Up, Down, or No (no differential expression) across five cancer cell lines \cite{yu2025vcworld}, jointly representing expression status and direction. We assess RL compatibility by applying GRPO to SFT/DFT/BFT initializations on 45 Tahoe-100M cell lines \cite{roohani2025tahoe100m}, with a binary correctness reward plus a format-validity term, and evaluating on the five held-out VCWorld cell lines. \textbf{General capabilities} are measured with 5-shot MMLU \cite{hendryckstest2021} and CMMLU \cite{li2024cmmlu}. \textbf{Representation tasks} cover gene-level prediction, cell-level clustering, multimodal integration, and perturbation prediction with STATE datasets \cite{adduri2025predicting}.

\subsection{Models and Baselines}

We employ the DeepSeek-R1-Distill series (14B Qwen, 32B Qwen, and 70B Llama) \cite{guo2025deepseek} as our base models; each size is independently fine-tuned on the same combined medical and biological data. We compare BFT against \textbf{SFT}, the standard negative log-likelihood objective; \textbf{DFT} \cite{wu2025generalization}, which scales each token loss by its confidence; and the \textbf{BFT-token} and \textbf{BFT-sample} ablations. BFT-token sets $s_b\equiv1$; BFT-sample sets $w_{b,t}^{\mathrm{BFT}}\equiv1$. For \textbf{agent pipelines} we keep the pipeline unchanged and replace only the backbone: GPT-4o in GeneAgent \cite{wang2025geneagent} and Gemini-2.5-Flash in VCWorld \cite{yu2025vcworld}. For representation tasks we compare against scGPT \cite{cui2024scgpt}, GenePT \cite{chen2025simple}, and STATE \cite{adduri2025predicting}.

SFT, DFT and BFT are trained with LlamaFactory \cite{zheng2024llamafactory} under a shared optimization recipe, differing only in the loss weights. GRPO uses TRL \texttt{GRPOTrainer} \cite{vonwerra2022trl}, with a shared hyperparameter grid to assess sensitivity to the default setting. Local open-weight runs use NVIDIA H20 GPUs; API workloads are separate. Appendix~\ref{app:training} gives optimization, reward, decoding and variance definitions. Appendices~\ref{app:families} and \ref{app:baselines} report additional model families and reweighting baselines.


\begin{figure}[t]
\centering
\includegraphics[width=\linewidth]{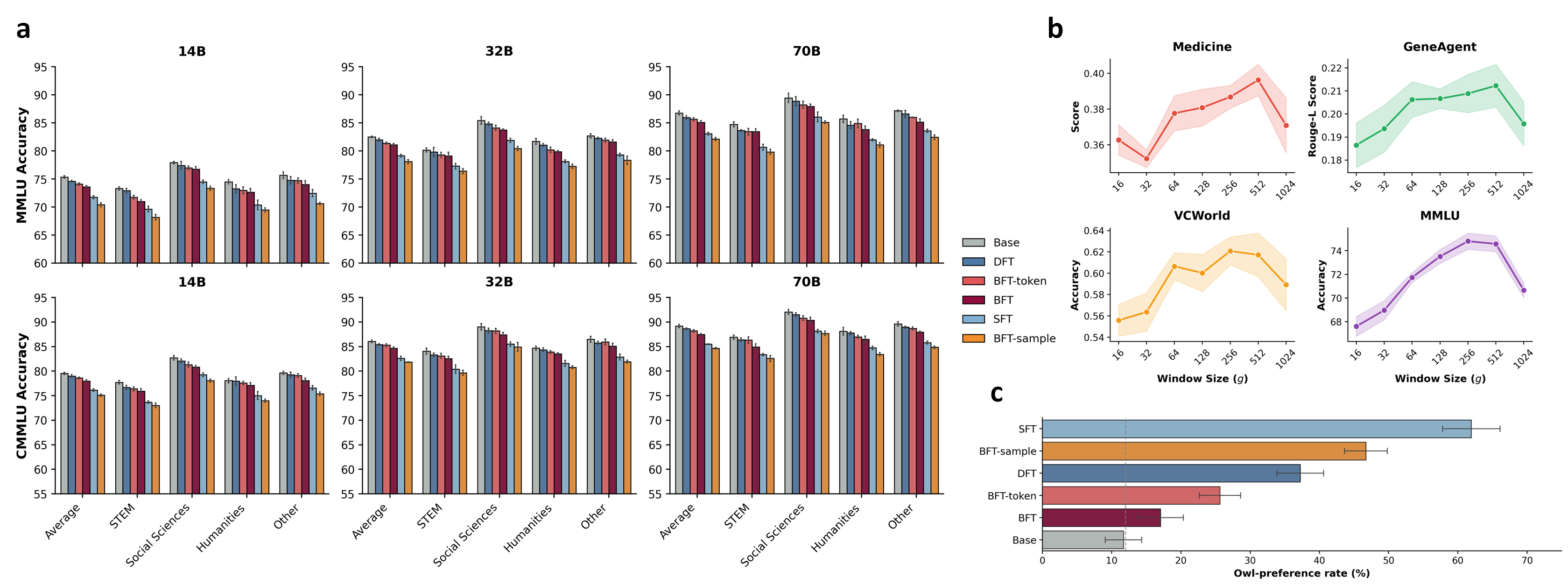}
\caption{
General capability retention, window sensitivity and preference transmission. \textbf{a}: MMLU and CMMLU evaluations. \textbf{b}: BFT performance under different window sizes $g$, measured on the 14B model; absolute values are therefore not comparable to the 70B results reported in Table~\ref{tab:vcworld}. \textbf{c}: Transmission of an injected owl preference.
}
\label{Figure3}
\end{figure}

\begin{table}[t]
\centering
\caption{Comparison of multimodal integration on the scIB benchmark \cite{luecken2022benchmarking}. Bio conservation (Isolated labels, KMeans NMI, cLISI) and Batch correction (Silhouette batch, iLISI, Graph connectivity) are reported per-metric; aggregate scores are reported separately, with Total = $0.4$\,Batch + $0.6$\,Bio.}
\label{tab:performance_comparison}
\resizebox{\linewidth}{!}{%
\begin{tabular}{l ccc ccc ccc}
\toprule
& \multicolumn{3}{c}{\textbf{Bio conservation}} & \multicolumn{3}{c}{\textbf{Batch correction}} & \multicolumn{3}{c}{\textbf{Aggregate score}} \\
\cmidrule(lr){2-4} \cmidrule(lr){5-7} \cmidrule(lr){8-10}
\textbf{Method} & \shortstack{Isolated\\labels} & \shortstack{KMeans\\NMI} & cLISI & \shortstack{Silhouette\\batch} & iLISI & \shortstack{Graph\\conn.} & \shortstack{Batch\\correction} & \shortstack{Bio\\conservation} & Total \\
\midrule
scMODAL & 0.54 & 0.66 & 1.00 & 0.87 & 0.46 & 0.69 & 0.62 & 0.67 & 0.65 \\
BFT     & 0.52 & 0.61 & 0.98 & 0.96 & 0.34 & 0.52 & 0.50 & 0.62 & 0.57 \\
Harmony & 0.48 & 0.44 & 0.98 & 0.75 & 0.08 & 0.48 & 0.38 & 0.58 & 0.50 \\
BBKNN   & 0.52 & 0.51 & 0.99 & 0.48 & 0.00 & 0.45 & 0.19 & 0.60 & 0.44 \\
\bottomrule
\end{tabular}%
}
\end{table}

\begin{figure}[t]
\centering
\includegraphics[width=\linewidth]{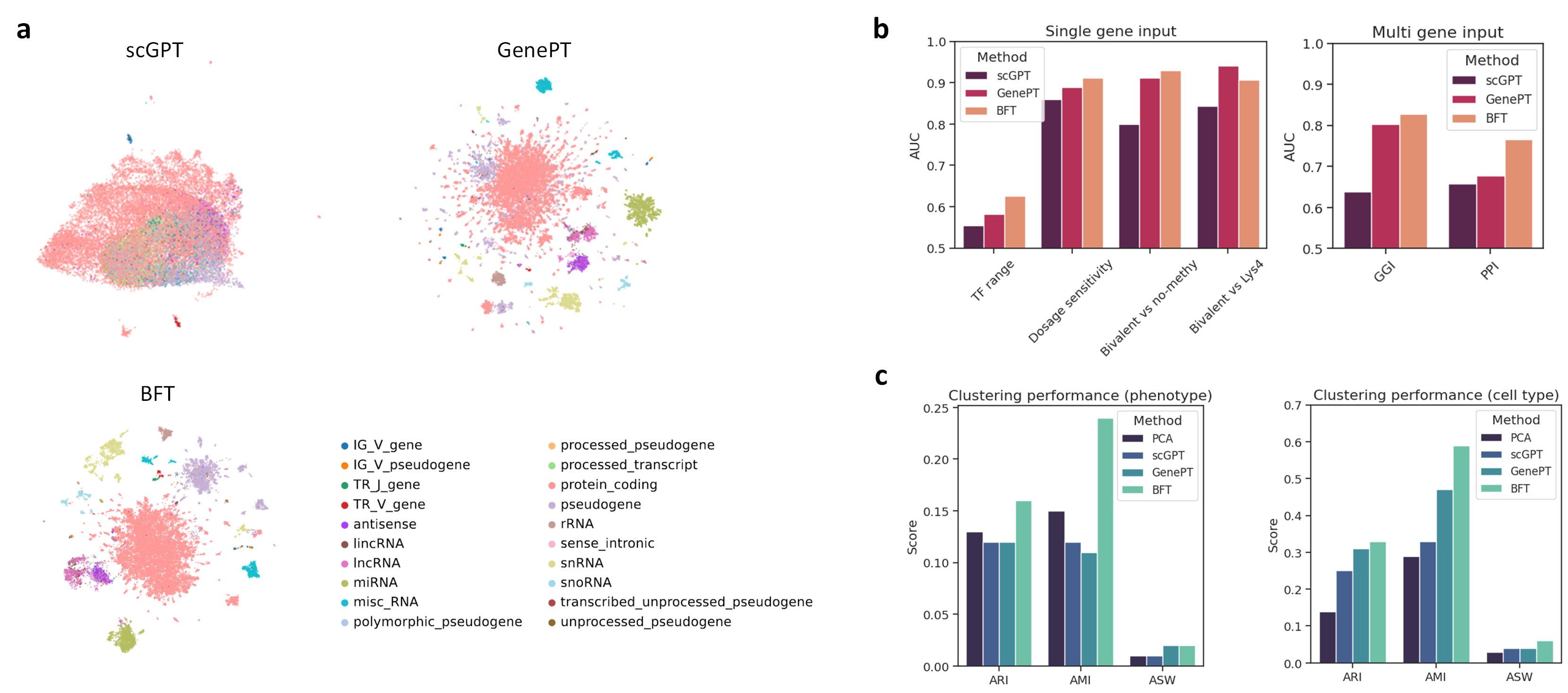}
\caption{
Unified representation across genetic and cellular levels. \textbf{a}: UMAP visualization of gene embeddings.
\textbf{b}: Gene-level downstream performance on biological property prediction and gene interaction prediction.
\textbf{c}: Cell-level representation performance on phenotype and cell-type.
}
\label{Figure4}
\end{figure}

\section{Results}

\subsection{Medical Competence, Biological Process and Chemical Perturbation Reasoning}

In medical competence, BFT improves more consistently than SFT, DFT, and the two ablated variants from 14B to 70B (Figure~\ref{Figure2}a). We further evaluate biological process reasoning using the GeneAgent benchmarks, where BFT improves with scale and achieves the best overall ROUGE among the post-trained open-weight models (Figure~\ref{Figure2}b).

BFT-token and BFT-sample provide intermediate gains, while SFT and DFT lag behind, suggesting that token-level group normalization and sequence-level reallocation are complementary. The claim is deliberately within recipe: BFT improves over the other post-trained open-weight models under the same agent pipelines, rather than uniformly exceeding every closed-source backbone or every subtask.

To further assess biomedical reasoning under a harder causal task, we evaluate on VCWorld \cite{yu2025vcworld}, replacing its Gemini-2.5-Flash backbone with SFT-, DFT-, and BFT-aligned DeepSeek-R1-Distill 70B models. BFT reaches 0.696 average accuracy (Table~\ref{tab:vcworld}). Across five cell lines it is numerically higher but not statistically distinguishable from Gemini-2.5-Flash ($+0.018$, 95\% CI $[-0.004,0.040]$, $p=0.088$), and exceeds the SFT and DFT replacements ($+0.134$, $p=0.0009$; $+0.206$, $p=0.0005$). These exploratory paired tests use the five cell-line means and are not adjusted for multiple comparisons; training-seed variability is reported separately in Table~\ref{tab:training-seeds}. DFT's lower score than SFT is consistent with, but does not by itself establish, suppression of knowledge-bearing low-confidence runs.

\subsection{BFT as a Robust Policy Initialization for Sparse-Reward RL}\label{sec:results-rl}

We next ask whether the post-trained models provide useful initializations for sparse-reward reinforcement learning. Using averages reported to three decimal places under the same GRPO setup, DFT and SFT decline from 0.490 to 0.420 and from 0.562 to 0.486, whereas BFT-sample, BFT-token, and BFT improve from 0.596 to 0.624, 0.660 to 0.694, and 0.696 to 0.736 (Table~\ref{tab:vcworld}). This opposing pattern is consistent with BFT providing a more RL-compatible initialization in this VCWorld setting. Automated analysis of 20 NSCLC trajectories finds more entities, causal connectives, and specific drug/mutation mentions under BFT; these are descriptive correlates rather than a causal explanation of the GRPO gain (Appendix~\ref{app:grpo_analysis}). Under a 256-token reasoning budget plus at most 32 completion tokens BFT still reaches 0.598 against SFT's 0.512, and the ordering holds at every tested budget (Appendix~\ref{app:length}). A separate KB check supports 81.4\% of BFT's length-matched, KB-alignable triples against 71.3\% for SFT, but only about one third of extracted triples are alignable and this is not a general factuality guarantee (Appendix~\ref{app:factuality}).

\subsection{General Capability Retention and Preference Transmission}

Beyond biomedical task performance, Figure~\ref{Figure3}a evaluates MMLU/CMMLU retention from 14B to 70B: DFT forgets least, followed by BFT-token and BFT, while SFT and BFT-sample degrade more. This ordering separates the roles of the two BFT components. The pattern is consistent with token weighting limiting forgetting and sample weighting emphasizing domain adaptation, but does not isolate the mechanism. Figure~\ref{Figure3}b is a separate 14B sensitivity study---its absolute values are not comparable to the 70B results in Table~\ref{tab:vcworld}---and evaluates sensitivity to $g$ across Medicine, GeneAgent, VCWorld, and MMLU. Performance depends on the window size, so continuous weighting does not eliminate this design choice. Figure~\ref{Figure3}c tests transmission of one injected owl preference \cite{cloud2026language}; BFT stays closest to Base and SFT transfers most strongly. This suggests that confidence-aware weighting can reduce the influence of a spurious pattern embedded in synthetic supervision, but the experiment covers one preference, one injection mechanism, and no clinical harm endpoint; it is therefore not a general biomedical-safety evaluation.

\begin{figure}[t]
\centering
\includegraphics[width=\linewidth]{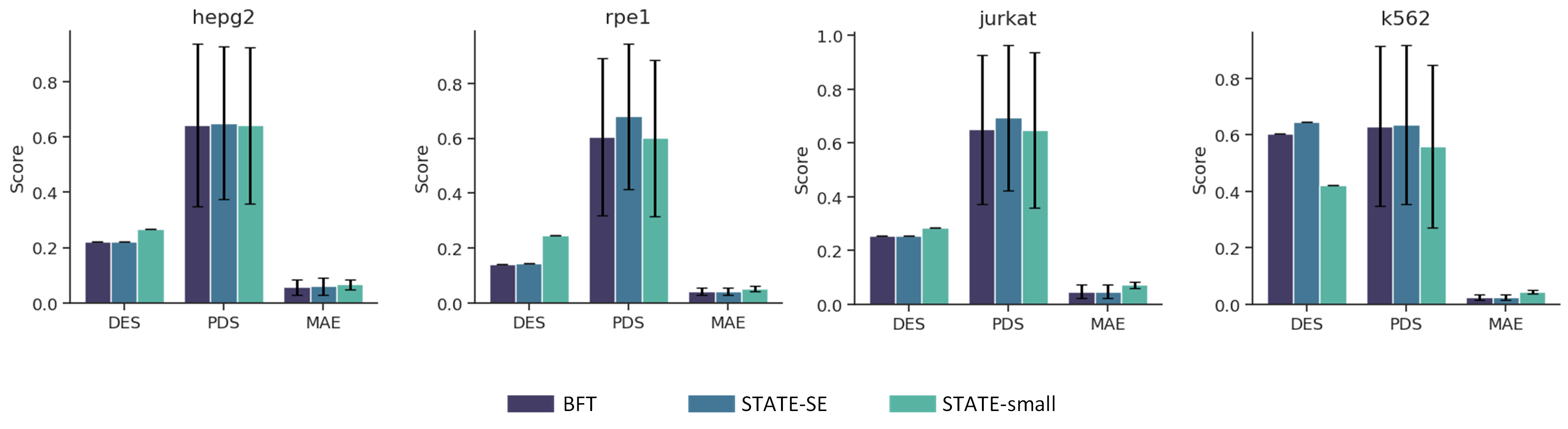}
\caption{Single-cell perturbation-response prediction across four datasets. BFT-generated text embeddings are used as inputs to the STATE decoder; profile generation is zero-shot with respect to the evaluated perturbation datasets, while the STATE decoder is retained.}
\label{Figure5}
\end{figure}

\subsection{Unified Representation Across Genetic and Cellular Levels}

We encode BFT-70B-generated entity profiles with \textsc{text-embedding-ada-002} \cite{neelakantan2022text}. GenePT uses the same encoder on NCBI summaries, so this comparison changes profile source and construction, not just the generating LLM. BFT gives the best average gene-level performance, although GenePT remains higher on Lys4 (Figure~\ref{Figure4}a--b; Table~\ref{tab:gene_representation}). Expression-weighted gene embeddings also outperform GenePT and scGPT on the evaluated cell-clustering tasks (Figure~\ref{Figure4}c; Table~\ref{tab:cell_clustering}). In multimodal integration, BFT reaches 0.57 overall scIB, below paired-supervision scMODAL (0.65) but above Harmony and BBKNN (Table~\ref{tab:performance_comparison}; Appendix~\ref{a-7}). With the STATE decoder, BFT embeddings perform comparably to standard inputs across four perturbation datasets (Figure~\ref{Figure5}) \cite{bunne2024build,adduri2025predicting,roohani2025virtual}. Only profile generation, not decoder training, is zero-shot with respect to these datasets.

\section{Conclusion}

We introduce Balanced Fine-Tuning (BFT), a confidence-aware objective motivated by the enrichment of biomedical concepts in dense low-confidence regions. BFT uses relative local confidence to preserve difficult tokens inside difficult contexts and a bounded sample coefficient to reallocate optimization toward sequences containing hard regions. The method remains close to ordinary supervised fine-tuning: it changes gradient allocation without requiring a reward model, an external uncertainty estimator, or diagnostic thresholds in the training loss. Across medical competence, biological reasoning, VCWorld, and subsequent GRPO, BFT improves over SFT and DFT under a shared recipe; its generated profiles also provide useful gene, cell, and multimodal representations. These findings support uncertainty structure as a practical biomedical post-training signal, but they have clear boundaries. The diagnostic partition is defined by base-model confidence, while training weights depend on the model's own confidence, so cross-family agreement and external annotation reduce but do not eliminate self-reference. Most supervision is source-conditioned synthetic text, and the targeted KB check covers only the alignable third of extracted triples. Finally, the GRPO, representation, and injected-preference results are specific to their evaluated pipelines and should not be read as general factuality, safety, or deployment guarantees. Unsupported biomedical outputs may mislead downstream research or clinical decisions, so consequential uses require independent verification. Extending BFT to independently calibrated confidence signals, naturally occurring expert supervision, and prospective biomedical evaluations is an important direction for future work.

\section*{Code availability}
Code is available at \url{https://github.com/TencentAILabHealthcare/BFT}.

\section*{Funding and competing interests}
This research received no funding. The authors declare no competing interests.

\bibliographystyle{unsrt}
\bibliography{reference}

\newpage
\appendix



\section{Appendix Texts}

\subsection{The Uncertainty Structure of Biomedical Text}\label{app:diag}

This section reports the full diagnostic behind the claim made in the Introduction: dense low-confidence regions in biomedical text are strongly enriched for domain concepts, whereas matched general-domain controls show no corresponding entity enrichment.

\paragraph{Casting each corpus as an instruction--response pair.}
The quantity the diagnostic measures, $c_t = \pi_\theta(y_t^*\mid y^*_{<t}, x)$, is the same quantity BFT reweights, so it is only meaningful relative to a prompt. Each of the four corpora in Table~\ref{tab:diag-corpora} is therefore evaluated in $(x,y^*)$ form. UltraChat, NuminaMath and our synthetic supervision have natural prompts. For PubMed Central full text, we construct $x$ from the article title and section heading and use the section body as $y^*$; this construction is stated explicitly rather than treated as a natural instruction. We keep passages of at least 1024 tokens and truncate at 4096. The descriptive diagnostic uses stride-one windows of length 256 to identify sparse- and dense-low regimes; BFT training instead assigns each target token the centered local context defined in Section~3. The group thresholds are used only for diagnosis and never enter the training objective.

\paragraph{Window classification.}
Writing $\tilde c$ for the median of $\{c_t\}$ within a response, a token is \emph{low-confidence} if $c_t < 0.5\,\tilde c$. For each window $W_t$ we record the low-confidence fraction $r_t$ and the longest contiguous low-confidence run $L_t$. Group~A is $r_t \in [0.05, 0.20]$ with $L_t \le 4$, and Group~B is $r_t \ge 0.30$ with $L_t \ge 16$. Windows satisfying neither condition are simply not counted; the two groups are the extremes of a continuum, not a partition of all windows. Every threshold here is descriptive. The BFT objective in Equation~\eqref{eq:bft_objective} reads none of them, depending only on $c_{b,t}$ and $C^{\mathrm{loc}}_{b,t}$, and the sensitivity of the diagnostic to these cuts is reported at the end of this section.

\begin{table}[H]
\centering
\caption{Matched diagnostic across real and synthetic biomedical text and two general-domain controls. Group-B prevalence measures how often dense-low windows occur. Entity A$\rightarrow$B is the within-corpus change in the share of biomedical entities among low-confidence tokens; this contrast, rather than an absolute cross-corpus entity rate, is our primary quantity. Causal-B is the share of Group-B low-confidence tokens matched by the fixed causal-connective list.}
\label{tab:diag-corpora}
\small
\resizebox{\linewidth}{!}{%
\begin{tabular}{llcccc}
\toprule
\textbf{Corpus} & \textbf{Domain / prompt} & \textbf{Group-B prev.} & \textbf{Entity A} & \textbf{Entity B} & \textbf{Causal-B} \\
\midrule
PubMed Central full text & biomedical, constructed $x$ & 20.4\% & 12.5\% & 60.9\% & 30.6\% \\
Our synthetic supervision & biomedical, natural $x$ & 18.6\% & 11.8\% & 57.4\% & 31.3\% \\
UltraChat-200k & conversational, natural $x$ & 2.9\% & 3.6\% & 4.1\% & 11.1\% \\
NuminaMath-CoT & mathematical, natural $x$ & 7.5\% & 3.0\% & 3.2\% & 42.6\% \\
\bottomrule
\end{tabular}
}
\end{table}

Three readings of Table~\ref{tab:diag-corpora} matter. The two biomedical corpora exhibit comparable entity shifts although one is human-written and the other machine-generated. NuminaMath-CoT is the informative negative control: it produces more dense-low windows and more causal connectives than conversational text, yet its entity share remains flat, so dense uncertainty alone does not imply enrichment for biomedical entities. The biomedical annotation scheme does not test whether mathematical concepts are enriched in mathematical text. UltraChat also shows almost no A$\rightarrow$B movement. Four additional controls test the main alternatives. Within the common 1024--2048-token band, Group-B entity share remains 58.7\% on biomedical text versus 3.7\% on general text. Stratified resampling by word-frequency decile and subword length reduces the biomedical--general gap in Group-B prevalence from $+15.2$ to $+10.7$ points but does not eliminate it, showing that rarity explains part, not all, of the effect. A temporally held-out recent-PMC subset gives 19.7\% Group-B prevalence, close to the main corpus's 20.4\%, and PubTator Central independently gives 57.2\% Group-B entity share, close to SciSpacy's 60.9\%. These controls address length, rarity, prior familiarity and tagger dependence separately.

Because stride-one windows overlap, we treat the window percentages as descriptive and do not regard windows as independent samples for significance testing. The CRAFT article-level control, cross-family location agreement, and downstream retention counterfactual provide independent forms of corroboration rather than token-level pseudo-replication.

\paragraph{Expert annotation.}
SciSpacy is itself a neural model trained on biomedical text, so its judgments and the base model's confidence could in principle be correlated through a shared training distribution rather than through the structure of the text. We therefore repeat the diagnostic on all 97 CRAFT articles \cite{bada2012concept}, which provide expert concept annotations. We align scored-token offsets to the corpus-provided character offsets and form a binary mask from the union of manual concept spans, counting overlapping annotations once; CRAFT labels never enter Group assignment. Of the low-confidence tokens inside dense windows, 63.8\% overlap a manually marked concept span, against 11.2\% inside sparse windows and 18.5\% over the corpus as a whole. This is external corroboration of concept enrichment, not direct ground truth for epistemic uncertainty or factual correctness.

\paragraph{Counterfactual: does the partition carry the effect?}
The measurements above are correlational. We therefore train discrete retention controls in which tokens selected by a fixed mask receive unit weight and all remaining valid tokens retain DFT's detached confidence weight. Selecting every token recovers SFT; the random mask is size-matched to the Group-B mask, and the Group-A and Group-B masks are computed once from the same pre-fine-tuning checkpoint. Each 70B variant is retrained from that checkpoint with the data, optimizer, and two-epoch schedule held fixed, then evaluated on VCWorld. Full BFT is shown separately because it uses continuous $\rho_{b,t}$ weights and the sample coefficient $s_b$ rather than a binary mask.

\begin{table}[H]
\centering
\caption{Retention counterfactual on VCWorld (70B). The four discrete controls differ only in which tokens receive unit rather than DFT weight; random selection is size-matched to Group~B. Full BFT is the continuous, threshold-free reference.}
\label{tab:counterfactual}
\small
\begin{tabular}{lc}
\toprule
\textbf{Tokens retained at full weight} & \textbf{VCWorld accuracy} \\
\midrule
All tokens (SFT) & 0.562 \\
Random selection, size-matched to dense windows & 0.571 \\
Sparse-low windows only & 0.508 \\
Dense-low windows only & \underline{0.674} \\
Full BFT (continuous) & \textbf{0.696} \\
\bottomrule
\end{tabular}
\end{table}

Random retention lands near SFT and sparse-region retention falls below it, while retaining the dense windows alone recovers most of the gain. This ordering is inconsistent with a pure ``retain more tokens'' account because the random condition retains exactly as many.

\paragraph{Robustness of the partition.}
Sweeping the low-confidence multiplier from 0.3 to 0.7 and the Group-B $(r_t,L_t)$ criterion over $(0.20,8)$, $(0.30,16)$ and $(0.40,32)$ gives fifteen combinations. The within-corpus entity-share shift on PubMed Central remains between $+32$ and $+59$ points, so no particular descriptive cutoff is load-bearing. Repeating the frozen diagnostic with DeepSeek-R1-Distill-70B, Qwen2.5-14B-Instruct and Llama-3.1-8B-Instruct gives Group-B prevalence/entity share of 20.4\%/60.9\%, 19.1\%/58.3\%, and 22.7\%/55.6\%, respectively. Their Group-B locations overlap at a positional IoU of 0.61--0.68, against 0.12 for randomly placed windows with the same marginal coverage. This cross-family agreement is evidence that the partition is not merely an idiosyncrasy of one model's confidence, although it does not remove self-reference from the training procedure.

\paragraph{What this does not establish.}
The partition is defined by the base model's own confidence and is then used to train that same model, and no control removes that circularity entirely. The semantic validation instead comes from human-annotated CRAFT spans, the partition reproduces across independently trained models, and the retention counterfactual is scored on an external benchmark rather than on the confidence signal itself. A base model that is systematically overconfident in some subdomain would produce no dense-low regions there, and BFT would offer that subdomain no protection; we have not characterized where that failure begins.

\subsection{Background: SFT and DFT Formulations}\label{t-1}

\paragraph{Supervised Fine-Tuning (SFT)}
Given a dataset $\mathcal{D} = \{(x, y^*)\}$ of instruction--response pairs, SFT minimizes the token-level cross-entropy loss
\begin{equation}
L_{\mathrm{SFT}}(\theta)
= \mathbb{E}_{(x, y^*) \sim \mathcal{D}}
\!\left[
- \sum_{t=1}^{|y^*|} \log \pi_\theta(y_t^* \mid y_{<t}^*, x)
\right],
\qquad
\ell_t := -\log \pi_\theta(y_t^* \mid y_{<t}^*, x),
\end{equation}
so that the per-token gradient is $-\nabla_\theta \log \pi_\theta(y_t^* \mid y_{<t}^*, x)$.

\textit{Token-level RL-equivalent form.} Because BFT and DFT are defined token-by-token, we work out the policy-gradient equivalent of SFT directly at the token level (rather than at the sequence level). The policy-gradient form of an RL objective $J(\theta) = \mathbb{E}_{x, y_t \sim \pi_\theta(\cdot \mid y_{<t}, x)}[r_t(x, y_{\le t})]$ is $\nabla_\theta J(\theta) = \mathbb{E}\big[\nabla_\theta \log \pi_\theta(y_t \mid y_{<t}, x)\, r_t(x, y_{\le t})\big]$. Applying importance sampling to the per-token gradient of $L_{\mathrm{SFT}}$ gives
\begin{equation}
- \nabla_\theta \log \pi_\theta(y_t^* \mid y_{<t}^*, x)
= \mathbb{E}_{y_t \sim \pi_\theta(\cdot \mid y_{<t}^*, x)}
\!\left[
\frac{\delta(y_t, y_t^*)}{\pi_\theta(y_t \mid y_{<t}^*, x)}
\big(-\nabla_\theta \log \pi_\theta(y_t \mid y_{<t}^*, x)\big)
\right],
\end{equation}
where $\delta(y_t, y_t^*)$ is the per-token Kronecker delta. Treating the implicit per-token reward as $r_{\mathrm{SFT},t}(x, y_{\le t}) = \delta(y_t, y_t^*)$ and the implicit per-token weight as $w_t(y_t \mid y_{<t}^*, x) = \pi_\theta(y_t \mid y_{<t}^*, x)^{-1}$, the SFT gradient at every position $t$ takes the RL-equivalent form
\begin{equation}
\begin{aligned}
- \nabla_\theta \log \pi_\theta(y_t^* \mid y_{<t}^*, x)
&= -\, \mathbb{E}_{y_t \sim \pi_\theta(\cdot \mid y_{<t}^*, x)}
\Big[
w_t(y_t \mid y_{<t}^*, x)\,r_{\mathrm{SFT},t}(x, y_{\le t})\\
&\hspace{7em}\cdot\nabla_\theta \log \pi_\theta(y_t \mid y_{<t}^*, x)
\Big].
\end{aligned}
\end{equation}
This representation makes the formal importance-weight contrast explicit at the same granularity as DFT and BFT: the \emph{RL-equivalent} per-token weight is $w_t = \pi_\theta(y_t^* \mid y_{<t}^*, x)^{-1} = c_t^{-1}$, which is unbounded as the reference-token confidence approaches zero. This does not mean that the ordinary softmax cross-entropy gradient with respect to the logits is itself unbounded; that gradient remains bounded. DFT and BFT can instead be read as different choices of \emph{detached} per-token loss weight on the same underlying token-level objective.

\paragraph{Dynamic Fine-Tuning (DFT)}
DFT \cite{wu2025generalization} stabilizes SFT by multiplying each token-level loss with its detached predicted probability:
\begin{equation}
L_{\mathrm{DFT}}(\theta)
=
\mathbb{E}_{(x, y^*) \sim \mathcal{D}}
\left[
\sum_{t=1}^{|y^*|}
\mathrm{sg}\!\big(c_t\big)
\, \ell_t
\right],
\end{equation}
where $c_t = \pi_\theta(y_t^* \mid y_{<t}^*, x)$ and $\mathrm{sg}(\cdot)$ is the stop-gradient operator. Composed with the token-level RL-equivalent form above, the effective importance-sampling weight in DFT becomes $w_t = c_t^{-1} \cdot c_t = 1$, removing the inverse-probability blow-up of SFT; the DFT training objective itself still multiplies each token loss by $\mathrm{sg}(c_t)$. However, this absolute-confidence rule suppresses tokens irrespective of whether their low confidence reflects sparse stylistic noise or dense biomedical content. BFT addresses this by replacing $c_t$ with the group-normalized weight $\rho_{b,t} = \mathrm{clip}(c_{b,t} / (C_{b,t}^{\mathrm{loc}} + \varepsilon), 0, 1)$ and by adding the sample-level coefficient $s_b$ at the sequence boundary.

\paragraph{Sequence-level normalization for SFT, DFT, and BFT}
For a fair comparison, all three baselines in our experiments share the same outer aggregation as the BFT objective in Equation~\eqref{eq:bft_objective}, namely a per-sample mean over valid tokens followed by a per-batch mean. Concretely, given a mini-batch $\{(x_b, y_b^*)\}_{b=1}^{B}$ with mask $m_{b,t}\in\{0,1\}$, we use
\begin{equation*}
L_{\mathrm{SFT}}(\theta)
=
\frac{1}{B}\sum_{b=1}^{B}
\frac{\sum_{t} m_{b,t}\, \ell_{b,t}}{\sum_{t} m_{b,t}+\varepsilon},
\qquad
L_{\mathrm{DFT}}(\theta)
=
\frac{1}{B}\sum_{b=1}^{B}
\frac{\sum_{t} m_{b,t}\, \mathrm{sg}(c_{b,t})\, \ell_{b,t}}{\sum_{t} m_{b,t}+\varepsilon},
\end{equation*}
which differ from the BFT objective in Equation~\eqref{eq:bft_objective} only in the per-token weight (none for SFT, $\mathrm{sg}(c_{b,t})$ for DFT, $\mathrm{sg}(\rho_{b,t})$ for BFT) and in the sample-level coefficient ($s_b$ for BFT, identically $1$ for SFT and DFT). All three methods use the same \emph{nominal} optimizer, learning-rate schedule, batch size, sequence length, and outer normalization. BFT differs through its confidence-derived token and sample weights; because the batch mean of $s_b$ can vary, these weights can also change the effective gradient scale.

\subsection{Data Construction, Splits and Leakage Control}\label{app:data}

\paragraph{Medical trajectories.}
HealthBench ships \textit{Consensus} (3{,}671 prompts) and \textit{Hard} (1{,}000 prompts) as separate files, but they are not disjoint: 586 prompt ids appear in both, that is 58.6\% of \textit{Hard}. Since we synthesize supervision from \textit{Consensus} and evaluate on \textit{Hard}, we remove the intersection by id first and generate candidates only from the remaining 3{,}085 prompts. GPT-5.1 receives the clinical instruction and its complete Consensus rubric and generates only the response trajectory. Sequential filtering retains 2{,}820 trajectories (91.4\%; mean length 1{,}842 tokens). The id removal precedes generation, so no \textit{Hard} evaluation prompt is shown to the synthesis model.

\paragraph{Biological instruction pairs.}
Source passages are the NCBI-derived gene and protein summaries distributed with GenePT. EasyDataset segments each passage and prompts GPT-OSS-120B for instruction--response pairs; the prompt supplies the passage, asks for questions answerable from it alone, and forbids reference to the passage as an artifact (``according to the text''). Sequential filtering retains 48{,}236 pairs (87.6\%; mean response length 1{,}127 tokens).

\paragraph{Filtering.}
For both sources we apply, in order, a 64-token minimum response length, MinHash near-duplicate removal at Jaccard similarity 0.85, rejection and disclaimer-template removal, and evaluation-overlap filtering. Biological responses additionally must contain at least one entity recognized by SciSpacy \texttt{en\_core\_sci\_lg}. Both sources are rubric- or passage-conditioned synthetic supervision rather than human gold annotation; source conditioning does not itself guarantee factual correctness.

\paragraph{Generation templates.}
The medical template has the fields \texttt{Input: \{clinical instruction; complete Consensus rubric\}}, \texttt{Task: produce one coherent response covering every rubric dimension}, and \texttt{Output: \{response only\}}. The biological template has \texttt{Source: \{NCBI-derived GenePT passage\}}, \texttt{Task: produce one instruction--response pair grounded only in the source}, and \texttt{Output: Instruction \{\ldots\}; Response \{\ldots\}}. The generators are instructed not to mention the supplied rubric or passage in their outputs.

\paragraph{Decontamination.}
Independently of the id-level removal above, every evaluation set is checked against the candidate training corpus by 13-gram overlap after lowercasing and punctuation stripping. This removes 0.00\% of the medical source and 0.03\% of the biological source, after which no evaluation 13-gram remains. We then form a source-stratified 95:5 train--validation split with seed 42. For the representation benchmarks the split is gene-disjoint: no gene whose profile is generated at evaluation time appears in any training instruction. We report these checks separately because id removal catches shared prompts, whereas the $n$-gram pass catches shared content reached through different prompts.

\paragraph{Mixture.}
After filtering, the combined dataset contains 2{,}820 medical trajectories and 48{,}236 biological instruction--response pairs, for 51{,}056 samples in total. The medical-to-biological mixture is approximately $1{:}17.1$ by sample count. The stratified 95:5 split places 48{,}503 examples in the supervised-training partition.

\paragraph{Control without instruction synthesis.}
To test whether the method merely fits teacher-generated response style, we remove the synthesis step and train directly on the underlying human-written NCBI passages as targets, matching the synthetic condition in example and token count. Absolute VCWorld accuracy is lower in this condition, but BFT still exceeds SFT by 0.118, close to the 0.134 gain with synthetic instruction data (Table~\ref{tab:raw-ncbi}).

\begin{table}[H]
\centering
\caption{VCWorld accuracy after supervised training directly on raw NCBI text, without instruction synthesis.}
\label{tab:raw-ncbi}
\small
\begin{tabular}{lc}
\toprule
\textbf{Method} & \textbf{VCWorld accuracy} \\
\midrule
DFT & 0.472 \\
SFT & 0.531 \\
BFT & \textbf{0.649} \\
\bottomrule
\end{tabular}
\end{table}

\subsection{Training and Evaluation Configuration}\label{app:training}

\paragraph{Supervised stage.}
SFT, DFT and BFT share one recipe and differ only in the confidence-derived loss weights described in Appendix~\ref{t-1}. We perform full-parameter fine-tuning for two epochs in LlamaFactory using AdamW, a learning rate of $1\times10^{-5}$, a cosine schedule with linear warmup over the first 3\% of steps, bf16 mixed precision and DeepSpeed ZeRO-3. The global batch size is 64 and the sequence-length cutoff is 4096. For the 70B runs, 32 NVIDIA H20 GPUs use one example per device and two gradient-accumulation steps. With 48{,}503 training examples, one epoch contains $\lceil48{,}503/64\rceil=758$ optimizer steps and the two-epoch schedule contains 1{,}516 steps. The window size is $g=256$ for all reported BFT runs; Figure~\ref{Figure3}b sweeps it. The numerical constant is $\varepsilon=10^{-8}$.

\paragraph{GRPO stage.}
GRPO is run for one epoch with the TRL \texttt{GRPOTrainer}, AdamW, bf16 and DeepSpeed ZeRO-3 on H20 GPUs. The shared default uses learning rate $5\times10^{-7}$, KL coefficient $\beta=0.05$, group size $G=8$, sampling temperature 0.7 and at most 2048 new tokens. Differential-expression status and direction are combined into a three-class target: Up, Down, or No (no differential expression). The task reward is one when the predicted class matches the Tahoe-100M-derived reference label and zero otherwise; a 0.1 format-validity reward is added when the output contains a parseable prediction. There is no length or process reward. We also apply the same $3\times3$ learning-rate--$\beta$ grid to every initialization, with $G$ fixed. Table~\ref{tab:vcworld} reports the shared default; grid search improves SFT and DFT to 0.517 and 0.447 but leaves both below their pre-GRPO values of 0.562 and 0.490, while BFT reaches 0.741 from 0.696. The reported grid therefore reduces dependence on one default, but does not rule out other optimization settings or establish a universal initialization ranking.

\paragraph{Interpreting post-GRPO declines.}
GRPO uses group-relative rewards on training prompts, not a guarantee of improvement on held-out cell lines. Finite-sample updates, the composite reward and distribution shift can yield lower test accuracy even if training reward increases. Table~\ref{tab:vcworld} reports held-out performance; it does not demonstrate a decrease in training reward, optimization collapse, or a specific mechanism of knowledge loss.

\paragraph{Decoding and variance.}
Unless stated otherwise, generation follows the VCWorld defaults (temperature 0.6). In Table~\ref{tab:vcworld} and Figures~\ref{Figure2}--\ref{Figure3}, error bars denote mean $\pm$ standard deviation over eight stochastic inference passes of one fine-tuned checkpoint on the same prompts; they quantify decoding variability only. In contrast, the count-valued entries in Table~\ref{tab:grpo_quality} are mean $\pm$ standard deviation across its 20 prompts after deterministic metric extraction. Fine-tuning variance is reported separately below and must not be inferred from either type of error bar.

\begin{table}[H]
\centering
\caption{VCWorld average accuracy across three independent fine-tuning seeds. The final column is the standard deviation across the three seed-level means, not decoding variability.}
\label{tab:training-seeds}
\small
\begin{tabular}{lcccc}
\toprule
\textbf{Method} & \textbf{Seed 1} & \textbf{Seed 2} & \textbf{Seed 3} & \textbf{SD} \\
\midrule
DFT 70B & 0.490 & 0.466 & 0.508 & 0.021 \\
SFT 70B & 0.562 & 0.541 & 0.578 & 0.019 \\
BFT 70B & 0.696 & 0.679 & 0.708 & 0.015 \\
\bottomrule
\end{tabular}
\end{table}
The three ranges are disjoint in these runs; the worst BFT seed (0.679) remains 0.101 above the best SFT seed (0.578). The paired tests in Section~5 use the five held-out cell-line means and therefore test consistency across cell lines rather than across only three training seeds.

\paragraph{Compute.}
Local training and evaluation of the open-weight models use NVIDIA H20 GPUs; third-party API calls for synthetic-data generation and benchmark baselines are separate. One 70B supervised post-training run takes approximately 36 hours on 32 H20 GPUs, or about 1{,}150 H20 GPU-hours; one 70B GRPO run uses approximately 770 H20 GPU-hours. These are per-run estimates rather than an audited total across every seed, ablation, hyperparameter setting, and evaluation. The diagnostic passes are inference-only and require one teacher-forcing pass per passage without gradient computation.

\paragraph{Sequence-length and mixture sensitivity.}
Shortening the training cutoff from 4096 to 2048 reduces BFT's VCWorld accuracy from 0.696 to 0.671. This dependence is expected because $s_b=1-\min_t C^{\mathrm{loc}}_{b,t}$ depends on how many local windows a sequence contains even when $g$ is fixed. Changing the medical:biological ratio from the default $1{:}17.1$ to $1{:}8$ and $1{:}1$ gives 0.691 and 0.677, respectively. The ordering remains useful across these settings, while the default long-context, biology-heavy mixture performs best.

\subsection{Controls for Sample-Weight Scale and Sequence Length}\label{app:sample-controls}

The minimum-based coefficient $s_b=1-\min_t C^{\mathrm{loc}}_{b,t}$ has two potential confounds. First, its batch mean changes the effective gradient scale. Second, a longer response exposes more candidate windows and therefore has more opportunity to attain a small minimum. We isolate these effects with two controls while leaving token weights, data, optimizer, and all other settings unchanged.

\paragraph{Batch-normalized sample weighting.}
We normalize the original coefficient to unit batch mean:
\begin{equation}
\widehat{s}_b
:=
\frac{s_b}
{\frac{1}{B}\sum_{j=1}^{B}s_j+\varepsilon}.
\end{equation}
This preserves relative allocation across samples while removing batch-to-batch variation in the mean sample weight.

\paragraph{Fixed-count local-window weighting.}
For every response, we choose the same $K=8$ approximately evenly spaced valid token positions, denoted $S_b$, and define
\begin{equation}
s_b^{(K)}
:=
\mathrm{sg}\!\left(
1-\min_{t\in S_b}C^{\mathrm{loc}}_{b,t}
\right),
\qquad |S_b|=K.
\end{equation}
Because every response contributes the same number of candidate local contexts, this control removes the direct extreme-value advantage of longer responses while retaining sensitivity to a difficult region. All filtered training responses contain at least 64 target tokens and therefore support eight distinct positions.

\begin{table}[H]
\centering
\caption{Sample-level controls on VCWorld (70B). All rows use the same supervised recipe. Best in each column is shown in \textbf{bold}.}
\label{tab:sample-controls}
\small
\begin{tabular}{lc}
\toprule
\textbf{Sample-level rule} & \textbf{VCWorld accuracy} \\
\midrule
No sample coefficient (BFT-token) & 0.660 \\
Original minimum coefficient & \underline{0.696} \\
Batch-normalized coefficient $\widehat{s}_b$ & \textbf{0.700} \\
Fixed-count coefficient $s_b^{(K)}$, $K=8$ & 0.690 \\
\bottomrule
\end{tabular}
\end{table}

Batch normalization slightly improves accuracy from 0.696 to 0.700, providing no evidence that BFT's gain depends on changing the mean gradient scale. Fixed-count sampling reaches 0.690, retaining 95.5\% of the original BFT--SFT gain and remaining 0.030 above BFT-token. Thus, the sample-level benefit persists when every response contributes the same number of candidate local contexts; sequence length may affect the raw minimum, but it does not account for the observed improvement.

\subsection{Additional Model Families}\label{app:families}

The main results use the DeepSeek-R1-Distill series. To check that the effect is not specific to one family or to distillation from one teacher, we repeat the supervised stage on two unrelated backbones under the same recipe and evaluate on VCWorld (Table~\ref{tab:families}).

\begin{table}[H]
\centering
\caption{VCWorld average accuracy across model families under an identical recipe. The DeepSeek-R1-Distill row is the 70B result from Table~\ref{tab:vcworld}; the other two families are smaller, so absolute values are lower and only the within-row ordering is meaningful.}
\label{tab:families}
\small
\begin{tabular}{lccc}
\toprule
\textbf{Base model} & \textbf{DFT} & \textbf{SFT} & \textbf{BFT} \\
\midrule
Qwen2.5-14B & 0.42 & 0.47 & \textbf{0.57} \\
Llama-3.1-8B & 0.37 & 0.42 & \textbf{0.52} \\
DeepSeek-R1-Distill-70B & 0.49 & 0.56 & \textbf{0.70} \\
\bottomrule
\end{tabular}
\end{table}

The ordering BFT $>$ SFT $>$ DFT holds in all three families, including the inversion in which DFT falls below plain SFT --- an ordering consistent with, but not proof of, the proposed suppression mechanism. The diagnostic itself also transfers without any training: as reported in Appendix~\ref{app:diag}, the Group~A/Group~B partition computed from these same three backbones agrees at a positional IoU of 0.61 to 0.68.

\subsection{Comparison with Other Reweighting Objectives}\label{app:baselines}

Section~\ref{sec:related} cites several recent token-level reweighting methods without evaluating them. We run four of them here under the recipe used for SFT, DFT and BFT: identical data, optimizer, schedule, sequence length and decoding, with only the objective changed (Table~\ref{tab:other-baselines}).

\begin{table}[H]
\centering
\caption{VCWorld average accuracy for alternative reweighting objectives, 70B, shared recipe.}
\label{tab:other-baselines}
\small
\begin{tabular}{lc}
\toprule
\textbf{Objective} & \textbf{VCWorld accuracy} \\
\midrule
DFT + BFT sample coefficient & 0.528 \\
iw-SFT \cite{qin2025supervised} & 0.578 \\
UFT \cite{liu2025uft} & 0.591 \\
Anchored SFT \cite{zhu2025anchored} & 0.607 \\
ProFit \cite{liu2026profit} & 0.618 \\
BFT (ours) & \textbf{0.696} \\
\bottomrule
\end{tabular}
\end{table}

The four published supervised reweighting baselines improve on plain SFT and all fall short of BFT; against the strongest, ProFit, BFT gains 0.078 across the five cell lines ($p=0.002$, exploratory paired test). Adding BFT's sample coefficient to DFT reaches only 0.528, showing that generic hard-sample emphasis does not recover the full method without local normalization. The published baselines reweight by importance, by anchoring to a reference policy, or by a fitness estimate, none of which distinguishes a low-confidence token in a confident neighbourhood from one inside an equally uncertain run. We exclude RAFT \cite{dong2023raft} because it requires an additional reward signal and cannot be placed under the same supervised-only recipe.

\subsection{GRPO Compatibility Analysis}\label{app:grpo_analysis}

We evaluated the response characteristics of the SFT-aligned, DFT-aligned, and BFT-aligned 70B models on 20 biomedical causal reasoning prompts covering NSCLC molecular mechanisms (e.g., KRAS G12C oncogenesis, KEAP1-NRF2 chemoresistance, PD-L1 immune evasion). Each model generated responses under the same decoding configuration (temperature 0.6, max tokens 2048).

\paragraph{Automated evaluation metrics.} To avoid an LLM-as-judge evaluation loop, we use reproducible lexical rules and deterministic SciSpacy entity recognition rather than generative-model judgments:

\begin{itemize}
\item \textbf{Biomedical entity count}: the number of unique biomedical mention spans detected by SciSpacy \cite{neumann-etal-2019-scispacy} (\texttt{en\_core\_sci\_lg} model) per response. This model detects untyped mentions; the count is not a validated count of specific entity classes.
\item \textbf{Causal connective count}: occurrences of causal and mechanistic language from a predefined list of 45 biomedical indicators, including action verbs (``activates'', ``inhibits'', ``phosphorylates'', ``upregulates''), consequence markers (``leads to'', ``results in'', ``confers resistance to''), and pathway connectives (``downstream of'', ``via'', ``through the \ldots\ pathway'').
\item \textbf{Response length}: total character count; length alone does not measure mechanistic depth.
\item \textbf{Drug/mutation mention rate}: fraction of responses that mention at least one specific drug name (matched against a curated list of 120 FDA-approved or clinical-stage oncology compounds) or genetic mutation (matched via regex pattern for standard HGVS-like notation, e.g., G12C, T790M, V600E).
\end{itemize}

\begin{table}[H]
\centering
\caption{Automated response analysis for SFT, DFT, and BFT-aligned 70B models on 20 biomedical causal reasoning prompts. Count-valued columns report mean $\pm$ standard deviation across prompts, not repeated inference runs. \textbf{Entities}: unique biomedical entities per response (SciSpacy \texttt{en\_core\_sci\_lg}). \textbf{Causal}: causal connective count from a 45-term indicator list. \textbf{Length}: response character count. \textbf{Drug/Mut.}: fraction of responses mentioning $\geq$1 specific drug or mutation. Metric extraction is deterministic conditional on the generated responses.}
\label{tab:grpo_quality}
\small
\begin{tabular}{lcccc}
\toprule
\textbf{Method} & \textbf{Entities} & \textbf{Causal} & \textbf{Length} & \textbf{Drug/Mut.} \\
\midrule
SFT 70B  & $32.4 \pm 8.2$  & $8.6 \pm 2.4$  & $3187 \pm 557$  & $75\%$ \\
DFT 70B  & $28.1 \pm 9.5$  & $7.2 \pm 2.8$  & $2756 \pm 549$  & $65\%$ \\
\textbf{BFT 70B} & $\mathbf{55.4} \pm 7.1$ & $\mathbf{15.8} \pm 3.1$ & $\mathbf{5458} \pm 635$ & $\mathbf{95\%}$ \\
\bottomrule
\end{tabular}
\end{table}

BFT produces responses with $71\%$ more biomedical entities than SFT and $97\%$ more than DFT, and its causal connective count nearly doubles ($+84\%$ over SFT, $+119\%$ over DFT).

These raw counts should not be read as greater entity \emph{density}. Per thousand characters, unique-entity counts are 10.17, 10.20, and 10.15 for SFT, DFT, and BFT, respectively; BFT responses are longer and contain more entities in absolute terms, not per unit of text. Two other metrics are less dependent on length. Causal connectives per thousand characters are 2.70 for SFT, 2.61 for DFT and 2.89 for BFT. The drug/mutation mention rate is a per-response binary: 95\% for BFT against 75\% and 65\%, although a longer response still has more opportunity to contain a mention, so the explicitly length-matched analyses below remain primary.

\paragraph{Rare entities and causal depth.}
We define rare entities as reference-answer entities in the lowest PubMed mention-frequency decile and measure whether they appear in the generated response. Recall is 0.41 for SFT and 0.63 for BFT, remaining 0.60 when BFT is length-matched to SFT's mean 3{,}187 characters. Automatically extracted dependency-path causal chains have mean depth 2.3, 2.1, and 3.6 for SFT, DFT, and BFT, respectively, and 3.3 for length-matched BFT. Together with the per-character connective rates above, these controls show that the ordering is not solely a consequence of BFT generating more characters, although they remain automated proxies rather than expert factuality judgments.

\paragraph{Implications for GRPO compatibility.} These response statistics describe differences between initializations, not their causal effect on RL. The GRPO task reward is outcome-based: it does not directly score intermediate entities or causal links. A richer response therefore need not receive a better reward. The observed association with subsequent improvement should not be interpreted as evidence of improved intermediate credit assignment.

\subsection{Length-Controlled Evaluation}\label{app:length}

BFT responses are longer than SFT and DFT responses (Table~\ref{tab:grpo_quality}), which raises the question of whether its advantage is a length effect rather than a content effect. This section tests that directly.

\paragraph{Protocol.}
A hard truncation may remove the final prediction needed for VCWorld scoring. We therefore use a reasoning budget with forced completion: generation stops at token $N$; if no prediction has been stated, a fixed answer-eliciting instruction is appended and at most 32 additional tokens are decoded. The budget and rule are shared across methods. Thus, the maximum generated output is $N+32$, not strictly $N$ tokens, and this control does not impose identical realized lengths.

The 2048-token row numerically matches the main results. Similar accuracy at 4096 tokens suggests limited aggregate sensitivity to increasing the cap; it does not establish that no responses were truncated. The completion instruction is an intervention whose effect may differ across methods.

\begin{table}[H]
\centering
\caption{VCWorld average accuracy (70B) under a shared reasoning budget $N$ and up to 32 additional forced-completion tokens. The 2048-token row matches Table~\ref{tab:vcworld}; 4096 tests sensitivity to a larger cap.}
\label{tab:length-control}
\small
\begin{tabular}{lccc}
\toprule
\textbf{Reasoning budget $N$} & \textbf{DFT} & \textbf{SFT} & \textbf{BFT} \\
\midrule
256 tokens  & 0.441 & 0.512 & \textbf{0.598} \\
512 tokens  & 0.462 & 0.537 & \textbf{0.641} \\
1024 tokens & 0.481 & 0.554 & \textbf{0.672} \\
2048 tokens (main setting) & 0.490 & 0.562 & \textbf{0.696} \\
4096 tokens & 0.491 & 0.564 & \textbf{0.698} \\
\bottomrule
\end{tabular}
\end{table}

At $N=256$ with up to 32 completion tokens, BFT leads SFT by 8.6 points and DFT by 15.7 points; the ordering holds at every budget (Table~\ref{tab:length-control}). BFT loses more than SFT when compressed ($-0.098$ versus $-0.050$ from the 2048-token setting), so part of its unconstrained gain accompanies longer reasoning. The 2048 and 4096 rows are nearly unchanged, indicating small aggregate sensitivity to increasing the cap. We therefore conclude only that the BFT advantage survives stringent length control, not that response length is irrelevant.

\paragraph{Representation side.}
The same question applies to the embedding results, where a longer gene profile could produce a richer embedding for reasons unrelated to biological content. We cap BFT profile generation at SFT's mean output length and re-encode the newly generated profiles with the same \textsc{text-embedding-ada-002}. Averaged over the six gene-level columns, BFT moves from 0.829 to 0.818, still above GenePT's 0.802. This is a generation-time control rather than post-hoc profile truncation.

\subsection{Targeted KB Support Check}\label{app:factuality}

We conduct a deliberately narrow automated support check on the same 20 NSCLC causal-reasoning trajectories used in Appendix~\ref{app:grpo_analysis}, one response per prompt per method. This analysis is intended to test whether the reported ordering reverses on readily checkable biomedical relations; it is not an exhaustive factuality evaluation and includes no human adjudication.

For the length-matched text-property control, we truncate each BFT response prefix to SFT's mean response length of 3{,}187 characters before extracting triples. Prefix truncation is appropriate for a metric computed on the retained text; it is not used for VCWorld accuracy, where removing the final prediction would invalidate scoring and the forced-completion protocol of Appendix~\ref{app:length} is used instead.

\paragraph{What the metric measures.}
Relation triples are extracted from each response, and a triple enters the checkable subset only when \emph{both} endpoints map to a SemMedDB concept or a DrugBank identifier. The reported rate is the fraction of that subset whose relation is supported by the knowledge base (Table~\ref{tab:factuality}). It is a support rate on an alignable subset, not a coverage figure over all claims, and it should not be read as one.

\begin{table}[H]
\centering
\caption{Targeted knowledge-base support check for extracted relation triples from 20 NSCLC trajectories. The length-matched row prefix-truncates BFT responses to SFT's mean 3{,}187-character length before extraction. Rates apply only to the KB-alignable subset.}
\label{tab:factuality}
\small
\begin{tabular}{lcccc}
\toprule
\textbf{Method} & \textbf{Triples} & \textbf{KB-alignable} & \textbf{Support rate} & \textbf{Adversarial reading} \\
\midrule
SFT 70B & 608 & 209 (34.4\%) & 71.3\% & 24.5\% \\
DFT 70B & 579 & 193 (33.3\%) & 68.9\% & 23.0\% \\
BFT 70B & 951 & 328 (34.5\%) & \textbf{82.6\%} & \textbf{28.5\%} \\
BFT 70B, length-matched & 641 & 215 (33.5\%) & 81.4\% & 27.3\% \\
\bottomrule
\end{tabular}
\end{table}

The alignable fraction is similar across methods (33.3--34.5\%), but similar coverage does not ensure comparable claim difficulty or eliminate subset-selection bias. Counting every unalignable triple as unsupported preserves the ordering in the last column; this sensitivity analysis is not an estimate of factual accuracy over all claims.

\paragraph{What falls outside, and why.}
Roughly two thirds of extracted triples are not alignable, and the attrition is on the entity side rather than the relation side: a triple's endpoints are open text spans, and one unmappable endpoint removes the whole triple. Assigning each excluded triple to the first applicable reason, about 61\% have at least one endpoint with no knowledge-base identifier at all (cell lines, assay and method names, compound codes, recently assigned gene symbols); about 24\% have an endpoint that is a composite process or state phrase rather than a single concept, such as ``loss of ERK phosphorylation''; and about 15\% have an endpoint that is a quantitative or experimental object such as a fold-change or an IC50 value. A triple whose endpoints both map but whose relation has no counterpart in the knowledge base is counted as unsupported rather than excluded, so the rates above are conservative in that respect.

This is the principal limitation of the check. SemMedDB is derived through the automatic SemRep pipeline, whereas DrugBank is curated; their role here is to provide a check independent of the SciSpacy-based descriptive analysis. The present analysis does not establish the factuality of the unalignable residue or of generated biomedical responses in general.


\subsection{More Gene-Level and Cell-Level Results}\label{b-4}

Results in Tables \ref{tab:gene_representation} and \ref{tab:cell_clustering} show that standard SFT and DFT remain relatively weak for this style of profile-derived biological representation, often underperforming scGPT and sometimes even PCA. DFT exhibits the largest degradation, consistent with the concern that suppressing low-confidence tokens can also remove reasoning chains and rare entities that matter for functional profile texts. In contrast, BFT achieves the best average gene-level score and the best evaluated clustering metrics (ARI, AMI, ASW), but not the best score on every gene-level task. The comparison against GenePT holds the OpenAI \textsc{text-embedding-ada-002} encoder fixed but contrasts GenePT's NCBI summaries with profiles generated by our BFT-aligned 70B model; source selection and profile construction also differ, so the difference cannot be attributed solely to the generating model. BFT 70B nevertheless matches or exceeds GenePT on all gene-level and cell-level metrics except Lys4. This advantage is also robust to the choice of encoder: under the smaller BioBERT encoder the absolute scores drop and BFT's relative gain over SFT shrinks (the experiment does not isolate encoder capacity, truncation, or pooling effects), but the ordering DFT $<$ SFT $<$ BFT remains unchanged on every column. Overall, these results are consistent with the view that BFT improves the specificity of generated biomedical profile texts in a way that benefits downstream representation learning.

\begin{table}[H]
\centering
\caption{Quantitative evaluation of biological representations at the gene level. We compare the performance of embeddings derived from different fine-tuning strategies (SFT, DFT, and BFT) across two text-embedding encoders. We use OpenAI \textsc{text-embedding-ada-002}, the encoder used by GenePT for NCBI gene summaries; BFT instead generates profile text before encoding. This comparison holds the encoder fixed, not the source text. BioBERT is included as a domain-specific encoder. All values are reported as AUC. Bold indicates the best performance in each column.}
\label{tab:gene_representation}
\small
\begin{tabular}{llcccccc}
\toprule
\textbf{Encoder} & \textbf{Method} & \textbf{TF Range} & \textbf{Dosage} & \textbf{No-methy} & \textbf{Lys4} & \textbf{GGI} & \textbf{PPI} \\ 
\midrule
- & scGPT & 0.557 & 0.853 & 0.804 & 0.847 & 0.635 & 0.661 \\
- & GenePT & 0.585 & 0.891 & 0.910 & \textbf{0.939} & 0.806 & 0.681 \\
\midrule
\multirow{3}{*}{ada-002} & DFT & 0.519 & 0.808 & 0.761 & 0.793 & 0.598 & 0.604 \\
 & SFT & 0.534 & 0.838 & 0.769 & 0.812 & 0.616 & 0.622 \\
 & \textbf{BFT (Ours)} & \textbf{0.629} & \textbf{0.913} & \textbf{0.927} & 0.910 & \textbf{0.823} & \textbf{0.769} \\
\midrule
\multirow{3}{*}{BioBERT} & DFT & 0.495 & 0.806 & 0.755 & 0.777 & 0.566 & 0.583 \\
 & SFT & 0.505 & 0.836 & 0.763 & 0.798 & 0.584 & 0.596 \\
 & \textbf{BFT (Ours)} & 0.591 & 0.872 & 0.879 & 0.864 & 0.741 & 0.707 \\
\bottomrule
\end{tabular}
\end{table}

\begin{table}[H]
\centering
\caption{Quantitative evaluation of cell-level clustering performance across phenotype and cell type labels. We compare embeddings derived from SFT, DFT, and BFT using \textsc{text-embedding-ada-002} (the encoder GenePT uses for NCBI summaries; our method encodes BFT-generated profiles) and BioBERT (a domain-specific encoder), against traditional baselines (PCA, scGPT, and GenePT). Bold indicates the best performance in each column.}
\label{tab:cell_clustering}
\small
\begin{tabular}{ll ccc ccc}
\toprule
 & & \multicolumn{3}{c}{\textbf{Phenotype Clustering}} & \multicolumn{3}{c}{\textbf{Cell Type Clustering}} \\
\cmidrule(lr){3-5} \cmidrule(lr){6-8}
\textbf{Encoder} & \textbf{Method} & \textbf{ARI}  & \textbf{AMI}  & \textbf{ASW}  & \textbf{ARI}  & \textbf{AMI}  & \textbf{ASW}  \\
\midrule
- & PCA & 0.14 & 0.16 & 0.011 & 0.15 & 0.28 & 0.031 \\
- & scGPT & 0.11 & 0.13 & 0.011 & 0.26 & 0.32 & 0.039 \\
- & GenePT & 0.13 & 0.12 & 0.019 & 0.30 & 0.46 & 0.041 \\
\midrule
\multirow{3}{*}{ada-002} & DFT & 0.09 & 0.11 & 0.003 & 0.10 & 0.21 & 0.016 \\
 & SFT & 0.11 & 0.13 & 0.006 & 0.12 & 0.25 & 0.021 \\
 & \textbf{BFT (Ours)} & \textbf{0.17} & \textbf{0.25} & \textbf{0.022} & \textbf{0.34} & \textbf{0.58} & \textbf{0.061} \\
\midrule
\multirow{3}{*}{BioBERT} & DFT & 0.07 & 0.09 & 0.002 & 0.08 & 0.18 & 0.012 \\
 & SFT & 0.09 & 0.11 & 0.005 & 0.10 & 0.22 & 0.018 \\
 & \textbf{BFT (Ours)} & 0.13 & 0.18 & 0.015 & 0.27 & 0.41 & 0.043 \\
\bottomrule
\end{tabular}
\end{table}

\newpage
\section{Appendix Figures}
\subsection{Workflow of Biological Embedding}\label{a-4}

\begin{figure}[h]
\centering
\begingroup
\setlength{\unitlength}{\linewidth}
\begin{picture}(1,0.460)
\put(0,0){\includegraphics[width=\linewidth]{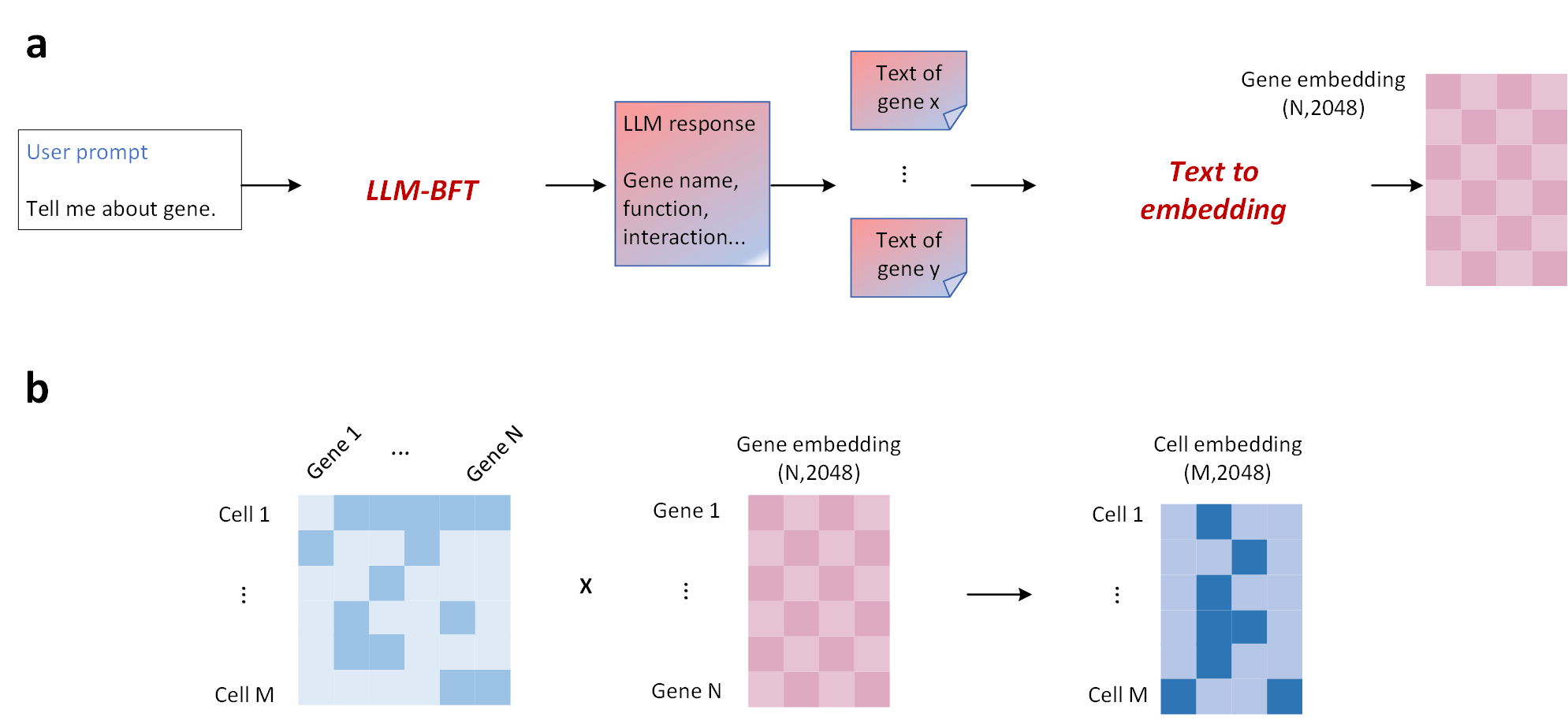}}
\put(0.812,0.380){\colorbox{white}{\fontsize{6}{7}\selectfont (N,1536)}}
\put(0.496,0.151){\colorbox{white}{\fontsize{6}{7}\selectfont (N,1536)}}
\put(0.756,0.151){\colorbox{white}{\fontsize{6}{7}\selectfont (M,1536)}}
\end{picture}
\endgroup
\caption{Workflow for extracting biological embeddings from LLM-BFT. \textbf{a}: LLM-BFT generates profile texts for entities of interest (e.g., a specific gene). The gene profile text is encoded by OpenAI \textsc{text-embedding-ada-002} to obtain gene embeddings. GenePT uses this encoder on NCBI gene summaries, whereas our method encodes BFT-70B-generated profiles. \textbf{b}: For a single-cell dataset, gene embeddings are weighted by gene expression values to generate cell embeddings.}
\label{ED-genept}
\end{figure}

\newpage
\subsection{UMAP of Multi-Gene Task}\label{a-5}

\begin{figure}[h]
\centering
\includegraphics[width=\linewidth]{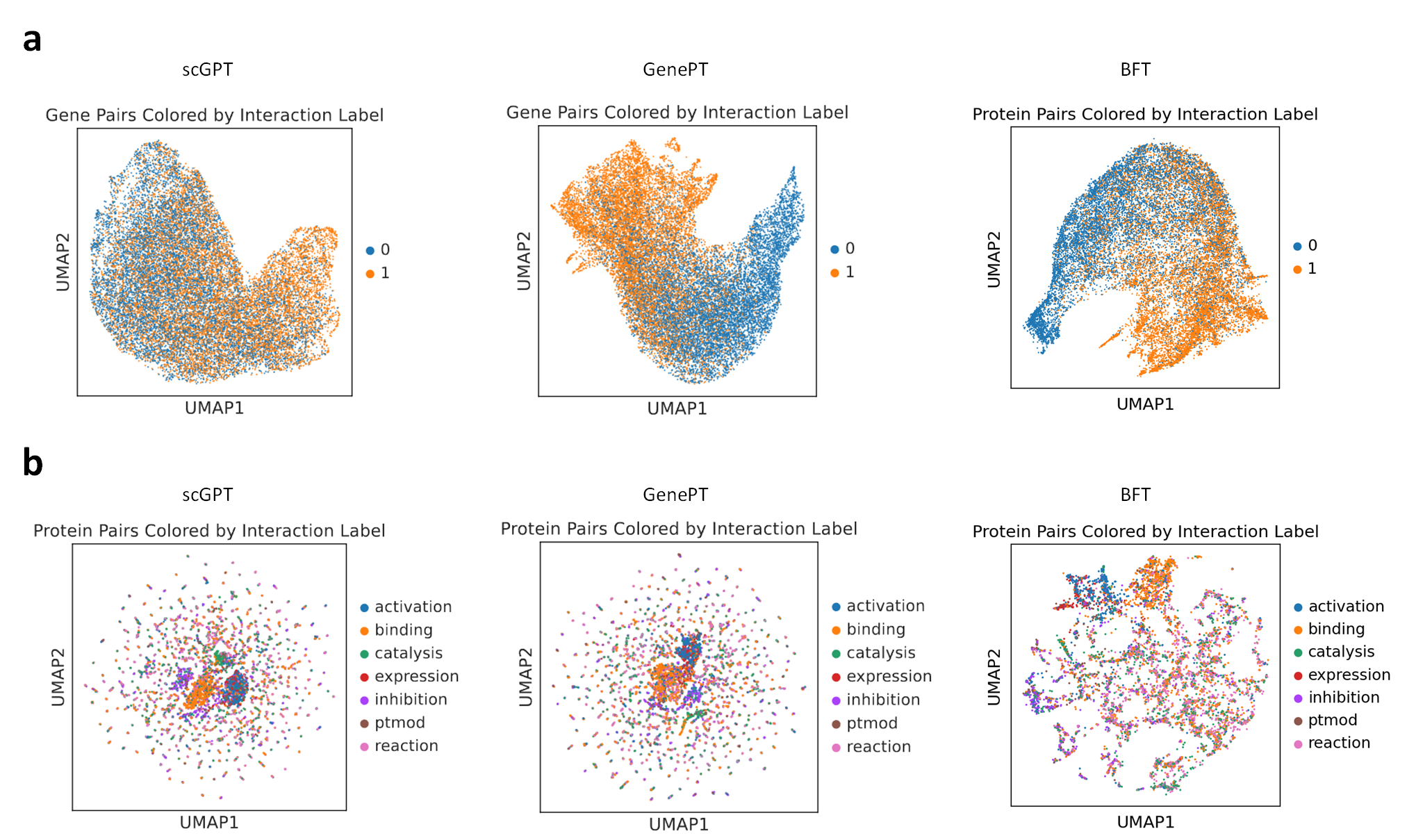}
\caption{UMAP visualization of the multi-gene input task. \textbf{a}: For GGI, the input embedding of the classifier is directly concatenated from the embeddings of two genes. \textbf{b}: For PPI, the input embedding of the classifier is directly concatenated from the embeddings of two proteins.}
\label{ED-gene-task}
\end{figure}

\newpage
\subsection{UMAP of Cell Embedding}\label{a-6}

\begin{figure}[h]
\centering
\includegraphics[width=\linewidth]{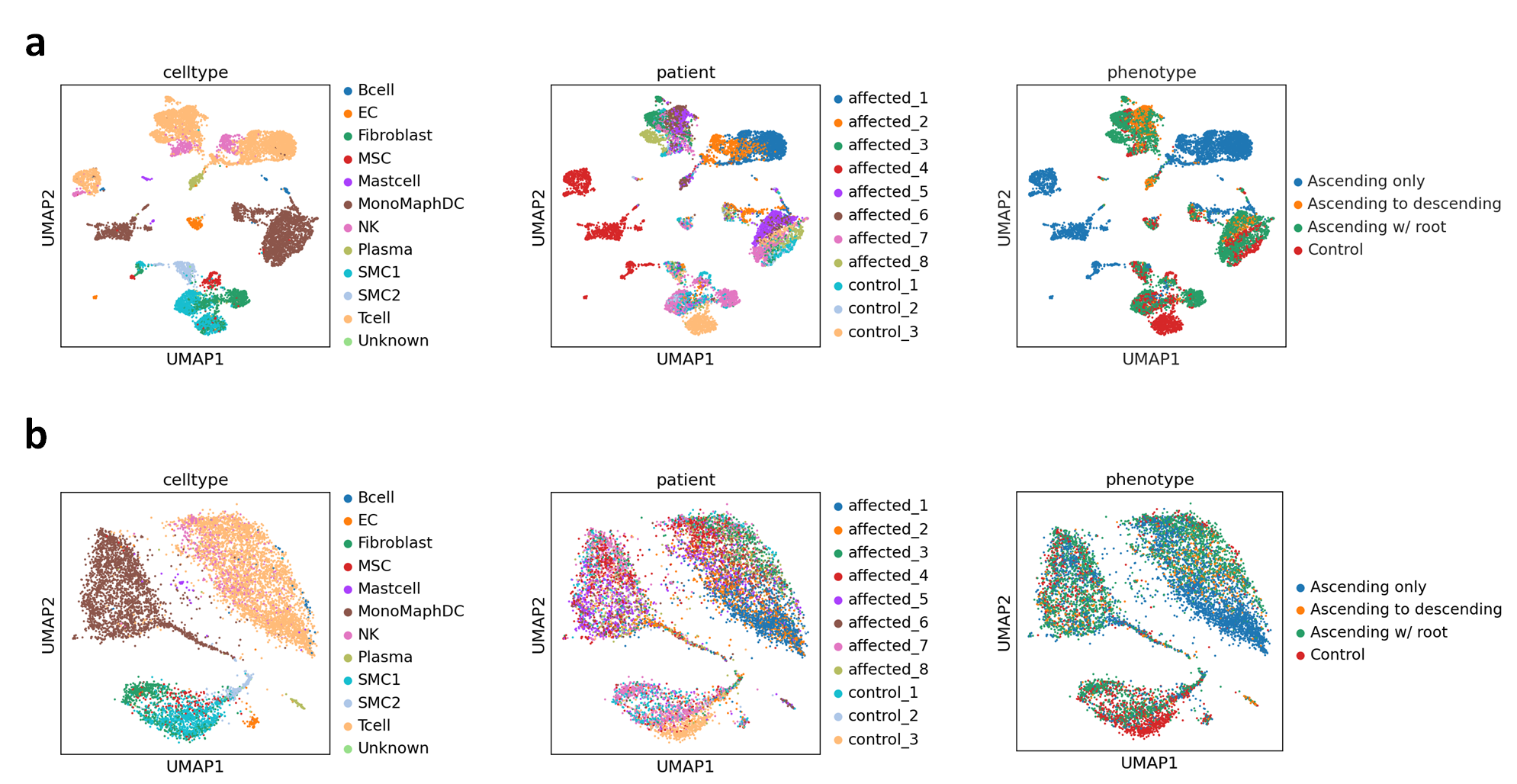}
\caption{UMAP visualization of cell-level embeddings. \textbf{a}: PCA embeddings of the raw data, colored by cell type labels (cell type heterogeneity), patient labels (batch labels), and phenotype labels (disease heterogeneity), respectively. \textbf{b}: Cell embeddings derived from LLM-BFT, colored by cell type labels (cell type heterogeneity), patient labels (batch labels), and phenotype labels (disease heterogeneity), respectively.}
\label{ED-cell-umap}
\end{figure}

\newpage
\subsection{UMAP of Multi-Modal Integration}\label{a-7}

\begin{figure}[h]
\centering
\includegraphics[width=\linewidth]{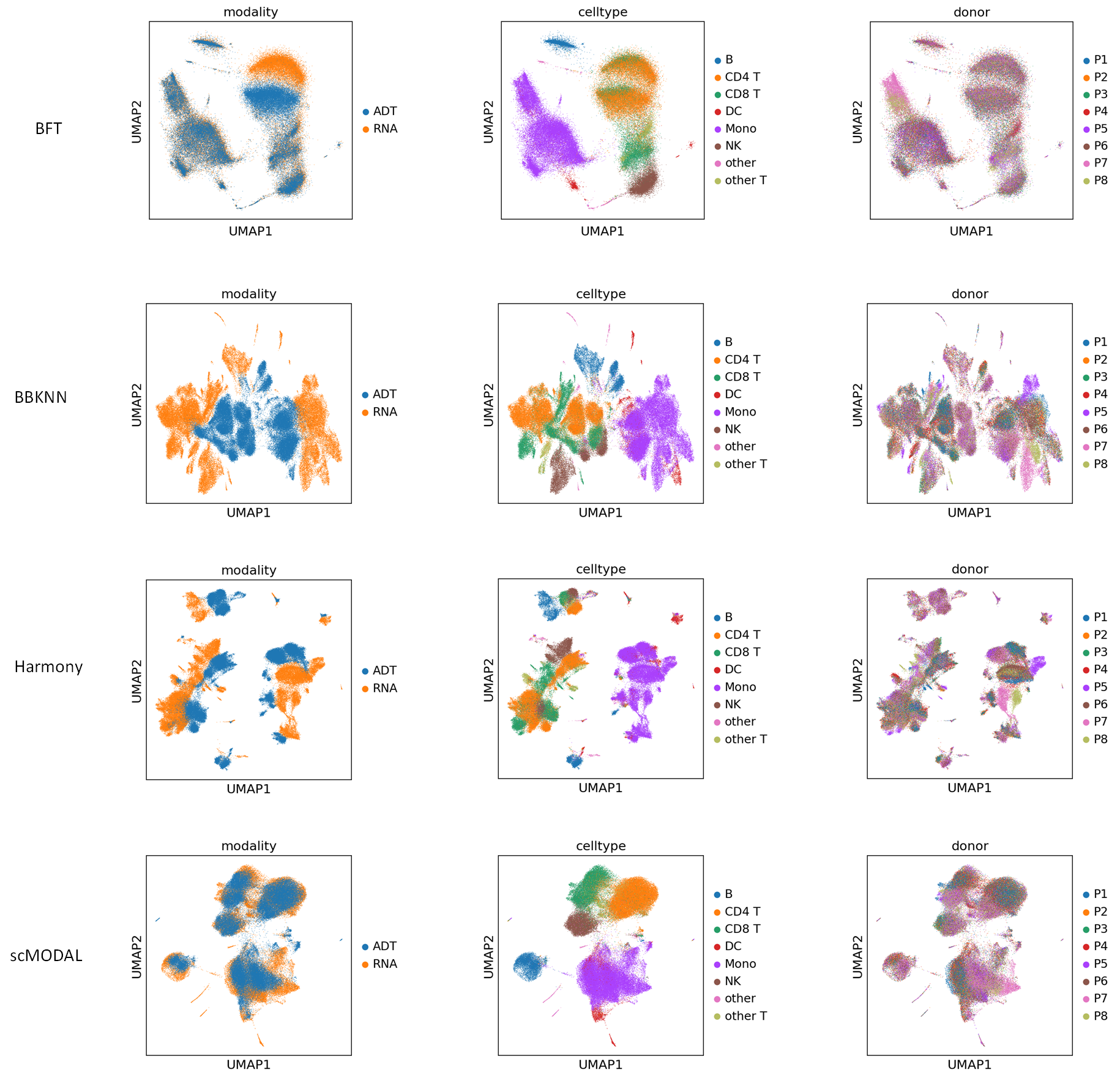}
\caption{UMAP visualization of single-cell multimodal data integration results. Rows 1 to 4 represent different integration methods, respectively. Columns 1 to 3 correspond to different coloring labels (modality, cell type, and donor), respectively.}
\label{ED-cell-integration}
\end{figure}



\end{document}